\documentclass[a4paper,12pt]{article}
\pdfoutput=1    

\usepackage[]{hyperref} 

\usepackage[utf8]{inputenc}
\usepackage{amsmath}
\usepackage{amsthm}
\usepackage{amssymb}
\usepackage{mathrsfs}
\usepackage{amsfonts}
\usepackage{bbold}

\usepackage{algorithm}
\usepackage{algorithmicx}
\usepackage{algpseudocode}
\usepackage{graphicx}
\usepackage{xifthen}
\usepackage{rotating}
\usepackage{xcolor}
\usepackage{subfig} 
\usepackage{psfrag}
\usepackage{amsopn}

\usepackage{authblk}

\usepackage{mathptmx}       
\usepackage{helvet}         
\usepackage{courier}        
\usepackage{type1cm}        
\usepackage{makeidx}         
\usepackage{graphicx}        
\usepackage{multicol}        
\usepackage[bottom]{footmisc}

\usepackage{ucs}              

\usepackage{latexsym}
\usepackage{amsmath,amssymb,amscd,amsbsy}
\usepackage{upgreek}

\usepackage{tensor}

\usepackage{makeidx}         
\usepackage{graphicx}        
\usepackage{multicol}        
\usepackage[bottom]{footmisc}

\usepackage{rcs} \RCS $Revision: 2.2 $ \RCS $Date: 2018/09/02 11:16:02 $ \RCS $Author: hgm $ \RCS $RCSfile: 18_RV-algebra-model.tex,v $ \RCS $Id: 18_RV-algebra-model.tex,v 2.2 2018/09/02 11:16:02 hgm Exp $

\usepackage{lmodern}   
\usepackage{anyfontsize}

\usepackage{mathrsfs}
\usepackage{bbm}
\usepackage{amsfonts}
\usepackage{amsmath}
\usepackage{enumerate}

\newcommand{\ignore}[1]{}

\newcommand{\vek}[1]{\mathchoice{\displaystyle\boldsymbol{#1}}
{\textstyle\boldsymbol{#1}}{\scriptstyle\boldsymbol{#1}}
{\scriptscriptstyle\boldsymbol{#1}}}

\newcommand{\ops}[1]{\mathchoice{\displaystyle\mathsf{#1}}
{\textstyle\mathsf{#1}}{\scriptstyle\mathsf{#1}}
{\scriptscriptstyle\mathsf{#1}}}

\newcommand{\tnb}[1]{\mathchoice{\displaystyle\mathboldsans{#1}}
{\textstyle\mathboldsans{#1}}{\scriptstyle\mathboldsans{#1}}
{\scriptscriptstyle\mathboldsans{#1}}}

\usepackage[sfdefault=cmbr,OMLmathsans]{isomath}  
\usepackage{latexsym}
\usepackage{amsmath}
\usepackage{amsthm}
\usepackage{amssymb}
\usepackage{upgreek}  
\usepackage{tensor}  

\usepackage[toc,title,page]{appendix}

\usepackage{dcolumn}
\usepackage{tikz}
\usetikzlibrary{shapes,arrows}
\usetikzlibrary{intersections}

\usepackage{float}
\usepackage{xcolor}

\makeindex             

\begin{document}

\title{An advanced spatio-temporal convolutional recurrent neural network for storm surge predictions
}

\author{{\small Ehsan Adeli$^{\dag}$, Luning Sun$^{\ddag}$, Jianxun Wang$^{\ddag}$, Alexandros A. Taflanidis$^{\dag}$}}
\affil{{\small $^{\dag}$ Department of Civil $\&$ Environmental Engineering $\&$ Earth Sciences\\ 
University of Notre Dame, Notre Dame, IN 46556, USA \\
$^{\ddag}$ Computational Mechanics $\&$ Scientific AI Lab, Department of Aerospace and Mechanical Engineering\\
University of Notre Dame, Notre Dame, IN 46556, USA \\ 
eadeli@nd.edu}}
\date{December 2021}

\maketitle

\abstract{In this research paper, we study the capability of artificial neural network models 
to emulate storm surge based on 
the storm track/size/\newline intensity history, leveraging a database of synthetic storm simulations. Traditionally, Computational Fluid Dynamics (CFD) solvers are employed to numerically solve the storm surge governing equations that are Partial Differential Equations (PDE) and are generally very costly to simulate. 
This study presents a neural network model that can predict storm surge, informed by a database of synthetic storm simulations. This model can 
serve as a fast and affordable 
emulator for the very expensive CFD solvers.
The neural network model is trained with the 
storm track parameters used to drive the CFD solvers, and the output of the model is the time-series evolution of the predicted storm surge across multiple nodes within the spatial domain of interest.
Once the model is trained, it can be deployed for further predictions based on 
new 
storm track inputs.
The developed neural network model is a time-series model, a Long short-term memory (LSTM), a variation of Recurrent Neural Network (RNN), which is enriched with Convolutional Neural Networks (CNNs). The convolutional neural network is employed to capture the correlation of data spatially (across the aforementioned nodes). Therefore, the temporal and spatial correlations of data are captured by the combination of the mentioned models, the ConvLSTM model. 
As the problem is a sequence to sequence time-series problem, an encoder-decoder ConvLSTM model is designed. Furthermore, the performance of the developed convolutional recurrent neural network model is improved by residual connection networks. Some other techniques in the process of model training are also employed to enrich the model performance so the model can learn from the data in a more effective way. 
The performance of the developed model is compared with the results provided by a Gaussian Process (GP) method, 
representing a popular alternative for establishing time-series emulation of storm surge predictions.
The results show the proposed 
convolutional recurrent neural network outperforms the GP implementation for the examined synthetic storm database.}

\section{Introduction}
\label{sec:Introduction}

Predicting future storm surge -related impact is receiving growing attention within the global scientific community, recognizing the widespread socio-economic implications of this natural hazard that need to be addressed within diverse prevention, mitigation, and post-disaster settings \cite{Hallegatte}.  Efforts to provide enhanced decision support against these imminent dangers over the past couple of decades have focused, among other topics, on numerical advances for storm surge predictions, producing high-fidelity simulation models that permit a detailed representation of hydrodynamic processes and therefore support high-accuracy forecasting. One such Computational Fluid Dynamics (CFD) solver, utilized later in this paper, is ADCIRC \cite{Luettich}, which is widely used \cite{Westerink} to simulate with high accuracy tidal circulation and storm surge propagation over large computational domains, and is, furthermore, typically coupled with appropriate models like SWAN \cite{Booij} or STWAVE \cite{Smith} to additionally incorporate wave action within the predictions. Unfortunately, the computational burden of such numerical models is large, requiring thousands of CPU hours for each simulation, something that limits their applicability for real-time surge forecasting (during landfalling events) or regional probabilistic flood studies. Due to this computational complexity, such models can be utilized to provide only a small number of high-fidelity, deterministic predictions, but cannot easily accommodate thousand-run storm ensembles, for example for examining the impact of forecast errors \cite{Kyprioti} in the predicted track during landfalling events. This dramatically limits their utility for decision makers either in emergency response management (during landfalling events) or regional planning (long-term projection of storm impact) settings.  \\

To address these computational challenges associated with high-fidelity solvers, and offer an alternative approach for probabilistic storm forecasting and risk assessment applications, machine learning tools and surrogate models have attracted significant attention \cite{Irish, Jia, Kim, Jia1, Kajbaf, Contento} for storm surge emulation. Based on databases of synthetic storm simulations, these approaches can provide fast-to-compute, data driven approximations for the expected storm surge. They are capable of replacing, with a high level of accuracy, the high-fidelity numerical model used that created the original database, maintaining the detailed underlying representation of hydrodynamic processes \cite{Resio}, while offering substantial computational efficiency. The latter efficiency makes them highly appropriate for supporting probabilistic surge forecasting and coastal hazard estimation applications. As such they can be leveraged to offer enhanced decision support for emergency response managers and regional planners \cite{Kijewski, Nadal}.\\

Among the different machine learning techniques that could be considered for this application, artificial neural networks has shown great promise \cite{Kim, Kajbaf, Lee3, Ramos}. This study, extends past efforts in this domain by considering a neural network implementation for predicting the entire time-series evolution of the storm surge using as input the time-series evolution of the storm track (latitude and longitude of eye of storm), intensity (pressure at center of storm) and size (radius of maximum winds). Past studies have focused on prediction of peak-surge only (as opposed to the time-evolution of the surge) and/or used instantaneous characteristics of the storm features as inputs for establishing the machine learning predictions. Should be pointed out that focus on prediction of peak-surge is common in most studies that have examined storm-surge emulations, with very few establishing predictions for the entire evolution of the storm surge. This study considers simultaneously time-series properties for both the surge predictions as well as the storm feature evolution, addressing, additionally, the spatial character of the predictions. To accommodate this substantial extension, a time-series Recurrent Neural Network model (RNN) is develop to predict the storm’s behavior. The spatial correlation of data, i.e the fact that the storm surge is estimated across multiple locations within the geographic domain of storm impact, is additionally considered by applying Convolutional Neural Networks (CNNs). Ultimately this allows both spatial and temporal correlations of data to be comprehensively captured by using a convolutional recurrent neural network model. The model input parameters to predict the storm surge are time-series for the storm track, size and intensity while the model output is the time-series of the storm surge level for specified locations along the coast.\\

Mathematically speaking, the typical neural network model maps the input parameters (layer) $\vek{z}_0 \in \mathbb{R}^{n_0}$ (the mentioned four input parameters) to the output $\vek{z}_L \in \mathbb{R}^{n_L}$ (surge values). The layers between the
input and output layers are the hidden layers $\vek{z}_l$, where $l = 1, ..., L$. Two adjacent layers are connected through the formulation below.

\begin{equation}
\vek{z}_l = \mathcal{F} (\tnb{W}^T_l \vek{z}_{l-1} + \vek{b}_l)
\label{eq:layers}
\end{equation} 

In Equation \ref{eq:layers}, $\tnb{W}$ and $ \vek{b}$ represents the model parameters, weight matrix and bias vector, respectively, and $\mathcal{F}$ denotes the activation function. After the model is trained, the model parameters are determined, and the output surge prediction can be rapidly computed from the given input parameters. This forward computation that involves only matrix multiplications 
has negligible computational burden compared to the original high-fidelity, CFD simulation. The model's performance is improved to capture more information from data through developing the applied neural network models. Other techniques to improve the model's efficiency are also considered in the training process. The results are compared to the results computed by a Gaussian process formulation \cite{Jia1}, which represents a state-of-the art alternative emulation technique for predicting time-series evolution of the storm surge. 
\\

In Section \ref{section:Storm Surge Prediction Problem Characteristics}, we discuss the 
problem formulation and the synthetic simulation data for training 
and testing for the machine learning models. Section \ref{section:Neural Network Methods} describes the machine learning methods and how the models are trained with the provided data. Then, results and comparisons that are followed by discussions are given in Section \ref{section:Model Evaluation}. Finally, the conclusions are given in Section \ref{section:Conclusions}.\\

\section{Storm Surge Prediction Problem Characteristics}
\label{section:Storm Surge Prediction Problem Characteristics}

The devastating flooding effects of numerous storms in the past two decades, such as hurricane Katrina and superstorm Sandy, have incentivized researchers to establish high-accuracy models to predict storm surge impact on coastal regions. These efforts have produced numerous advanced numerical CFD solvers \cite{Resio, Westerink, Jelesnianski, Luettich} used by various actors for emergency response management or regional planning. These solvers simulate the storm surge by solving the shallow water wave equations given the initial and boundary conditions. The simulation is driven by the atmospheric pressure and wind velocity that describes the time evolution of the hurricane vortex. This wind and velocity input can be derived through information for the storm track (location of center of rotation and forward speed of the vortex), size and intensity \cite{Holland, Holland1}, with intensity described by the wind speed or the pressure loss between the center and the far-away ambient conditions and size by the distance between the center and the location of maximum wind speeds. Interested readers can found additional information for hurricane physics and modeling in \cite{Marks}. These numerical tools can be ultimately used to accommodate deterministic and probabilistic approaches for establishing storm surge predictions \cite{Liberto, Dresback, Davis, Bernier, Kyprioti, Taylor}. \\

As discussed in the introduction, the aforementioned models provide high-accuracy estimates (empowered by high resolution spatial grids), but entail a very large computational cost that posed a great challenge for their widespread use, especially in the context of probabilistic assessments for real-time forecasting applications. To overcome this challenge, machine learning techniques can be developed that leverage precomputed datasets of synthetic hurricane simulations, providing information for storm parameters, paths and surge responses. Within this setting, the unknown functional relationship between inputs (hurricane parameters) and responses (storm surge) can be approximated by some type of regression, response surface or non-parametric emulation model. Specifically, this study focuses on Artificial Neural Network (ANNs) implementation. Substantial research efforts have already been made to consider ANN applications within storm surge emulation setting. \\

Lee et al. \cite{Lee, Lee1, Lee2} conducted research on shallow networks with a limited number of neurons to predict the storm surge for a few typhoons impacting Taiwan. A similar study with almost a similar size of networks has been carried out by De Oliveira et al. \cite{Oliveira} for the southeast coastal region of Brazil. Another study to reduce the uncertainty of storm surge prediction for Venice, Italy, is conducted by Bajo et al. \cite{Bajo}, 
again using shallow neural networks. It should be pointed out that in the mentioned studies, the neural network models are trained with very few storms, 
limiting predictive potential of the network and ability to establish in-depth learning form the data. 
To improve the model's performance, Kim et al. \cite{Kim} has used a bigger set of data established by using the ADCIRC model for the New Orleans region, and trained a shallow network which is tested on historical hurricane Katrina. Note that the model they applied was not a time-series neural network model, and therefore the temporal correlation of data was not properly leveraged within the model development. Several other similar studies have been carried out by Hashemi et al. \cite{Hashemi}, Kim et al. \cite{Kim1}, Chao et al. \cite{Chao}  and Das et al. \cite{Das} for other geographical regions, using larger training datasets (with larger number of storms) to train neural network models. However, these efforts did not, once again, consider the temporal correlation of the storm data. \\ 

More recently, a number of studies have employed time-series models to predict storm surge based on the time-evolution of the storm input parameters, in all cases utilizing a small number of storm simulations. Alemany et al. \cite{Alemany} has employed a recurrent neural network (RNN) to predict the storm surge when it gets close to the beach based on the very initial part of the surge. Igarashi et al. \cite{Chen} has also employed a standard recurrent neural network by utilizing a database of about 150 storm datasets to estimate the surge for upcoming storms. Furthermore, Chen et al. \cite{Igarashi} have applied a standard modification of the time-series model called Long Short-Term Memory (LSTM) model and trained it with a database of twelve storms.  \\

In most of the mentioned studies above, the models are trained with a limited number of storm observations. Also in all these studies, the number of grid points for which the surge is predicted is small. This is accomplished either by examining a small geographic region only, or by establishing some type of clustering approach, to reduce the original grid to a smaller number of representative points. Moreover, the standard sequence neural network models are mostly used as a black box in these studies, and no development and further investigation is applied to the standard time-series models, to accommodate some of the unique features of the storm surge emulation problem. \\

\begin{figure}[H]
    \begin{center}{}
      \includegraphics[width=3.75in]{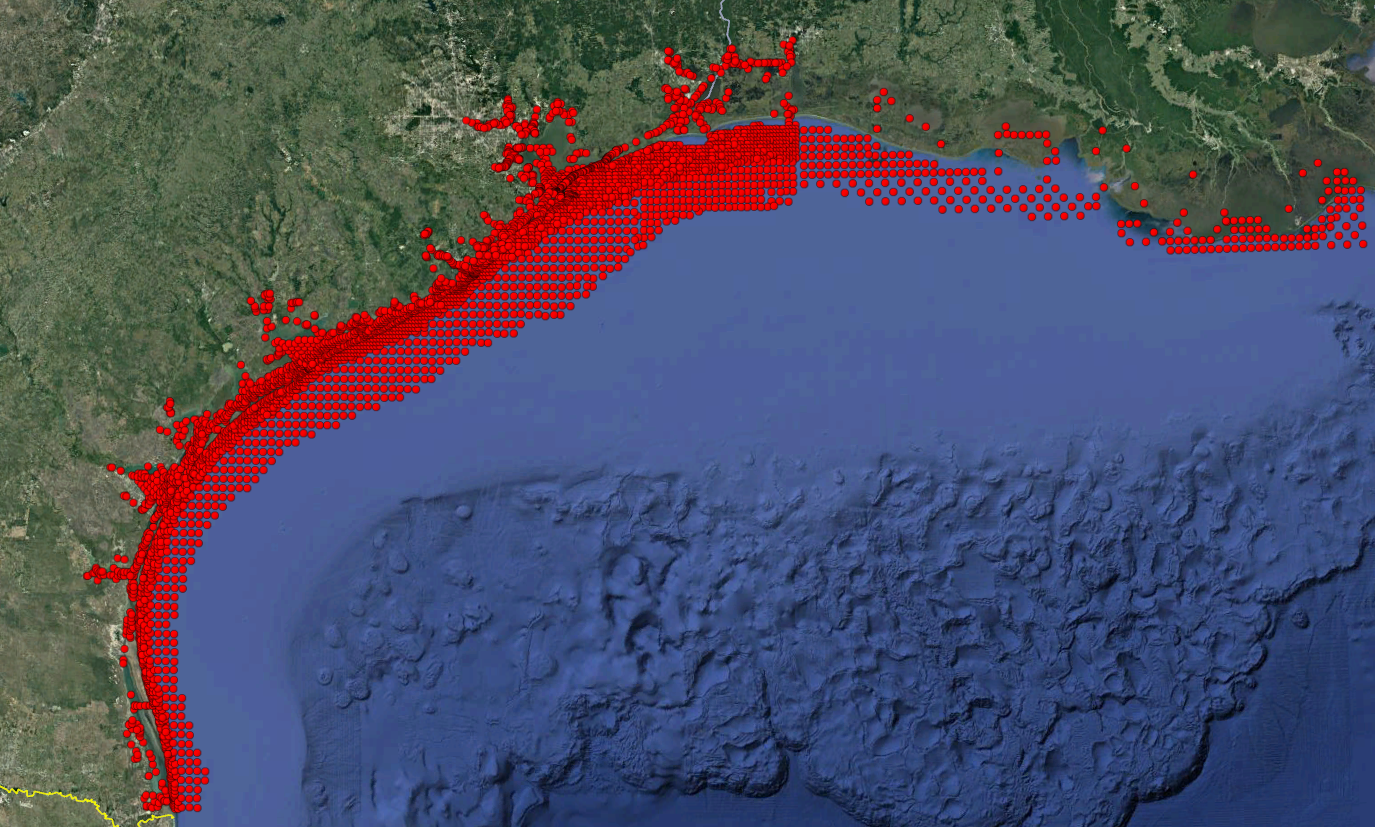}
      \caption{\label{fig:Alwet1} Grid of save points within the region of study.}
    \end{center}\hspace{2pc}%
\end{figure}

This study, extends these past efforts and considers a neural network implementation for predicting the time-series evolution of the storm surge across a geographic domain including a large number of save points (SPs), utilizing a database with a large number of storm surge simulations. The database is part of the Army Corps of Engineers Coastal Hazard System \cite{Nadal} and corresponds to synthetic storms simulations for the greater Coastal Texas region with a total of 4800 SPs, also shown in Figure \ref{fig:Alwet1}. 500 storms will be used for calibration of the neural network emulator and an additional 8 storms will be used as test-sample for its validation. The input for the synthetic storm simulations corresponds to: the latitude and longitude of the storm center (storm track parameters), the central pressure deficit (storm intensity parameter) and the radius of max winds (storm size parameter). The time-evolution for all four these parameters is utilized as input to the neural network. Note that some recent studies have considered some additional, derived parameters for describing the neural network input, namely the forward speed and the track heading \cite{Lee3}, but these correspond to redundant storm characteristics if time-evolution of the storm features is examined (instead of instantaneous features), and contribute to over-parameterization of the database. As such the input is represented by only 4 storm parameters. The predicted output corresponds to the storm surge across the 4800 SPs. 
This creates a sequence-to-sequence prediction problem, with both the input and the output of the neural network corresponding to sequences. Such type of problems are widely acknowledged to be exceptional challenging sequence prediction problems. \\

\begin{figure}[H]
    \begin{center}{}
      \includegraphics[width=2.25in]{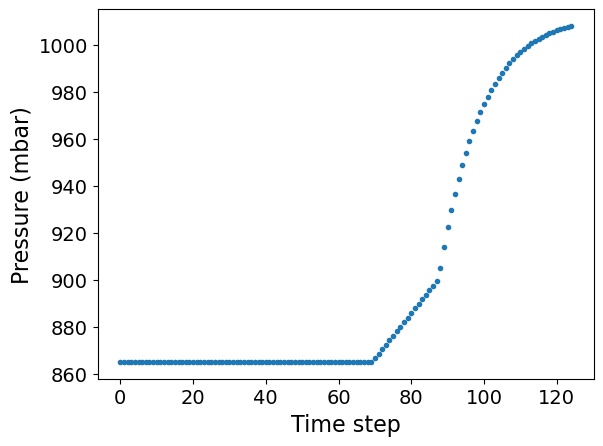}  
      \includegraphics[width=2.25in]{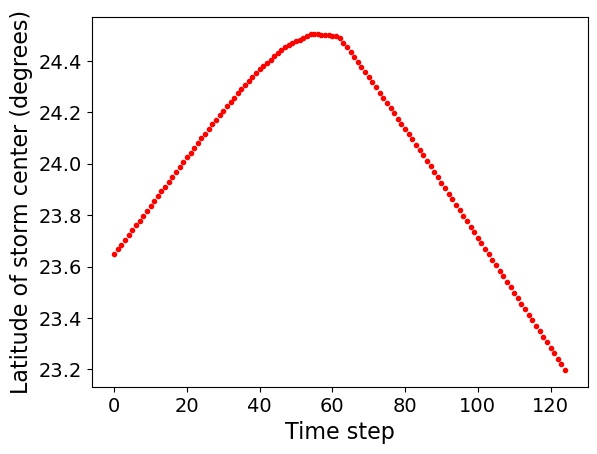}
      \includegraphics[width=2.4in]{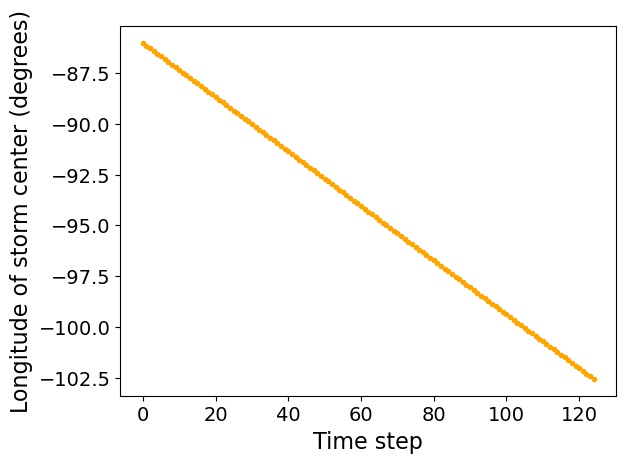}  
      \includegraphics[width=2.25in]{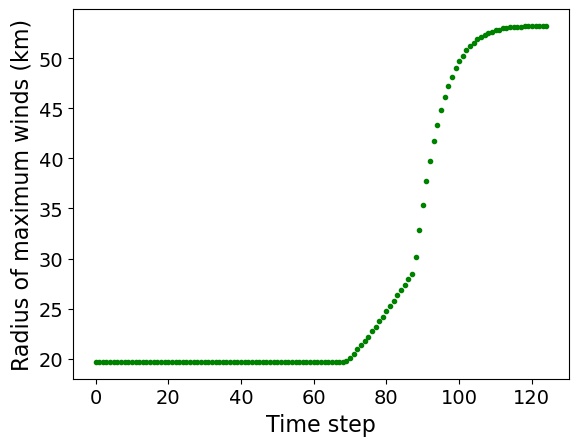}    
      \caption{\label{fig:inputs0} Input parameters for a typical database storm.}
    \end{center}\hspace{2pc}%
\end{figure}

\begin{figure}[H]
    \begin{center}{}
      \includegraphics[width=3.0in]{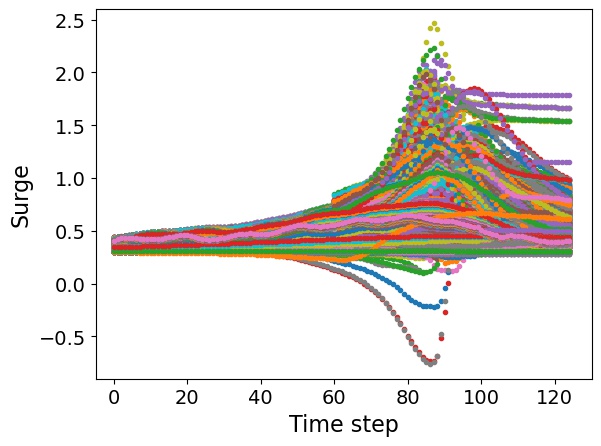}    
      \caption{\label{fig:output200} Surge output for different SPs for a typical database storm.}
    \end{center}\hspace{2pc}%
\end{figure}

For both the input and the output, 125 time steps are utilized, extending from the time each storm is couple thousands of kilometers before making landfall, to few hundred kilometers after making landfall. This range is chosen to encompass the time instances the maximum surge manifests across the entire geographic domain of interest. Synchronization of the time-series is established with respect to the landfall for each storm, as done in past studies \cite{Kim}. This landfall corresponds roughly to step 90. Figures \ref{fig:inputs0} and \ref{fig:output200} show variation of the 
four input parameters for a typical storm and the variation of the surge for different nodes for the same storm, respectively. 
It is evident from this figure that the size and intensity of the synthetic storms remain practically unchanged before the storm makes landfall.  This is common characteristic of many synthetic storm databases and creates some challenges for the neural network application as will be detailed later. 

In the following section, we thoroughly discuss the model and the applied developments of the model structure with the additional techniques.

\section{Neural Network Methods}
\label{section:Neural Network Methods}

\subsection{Convolutional Long short-term memory}
\label{subsection:Convolutional Long short-term memory}

\begin{figure}[H]
    \begin{center}{}
      \includegraphics[width=4.5in]{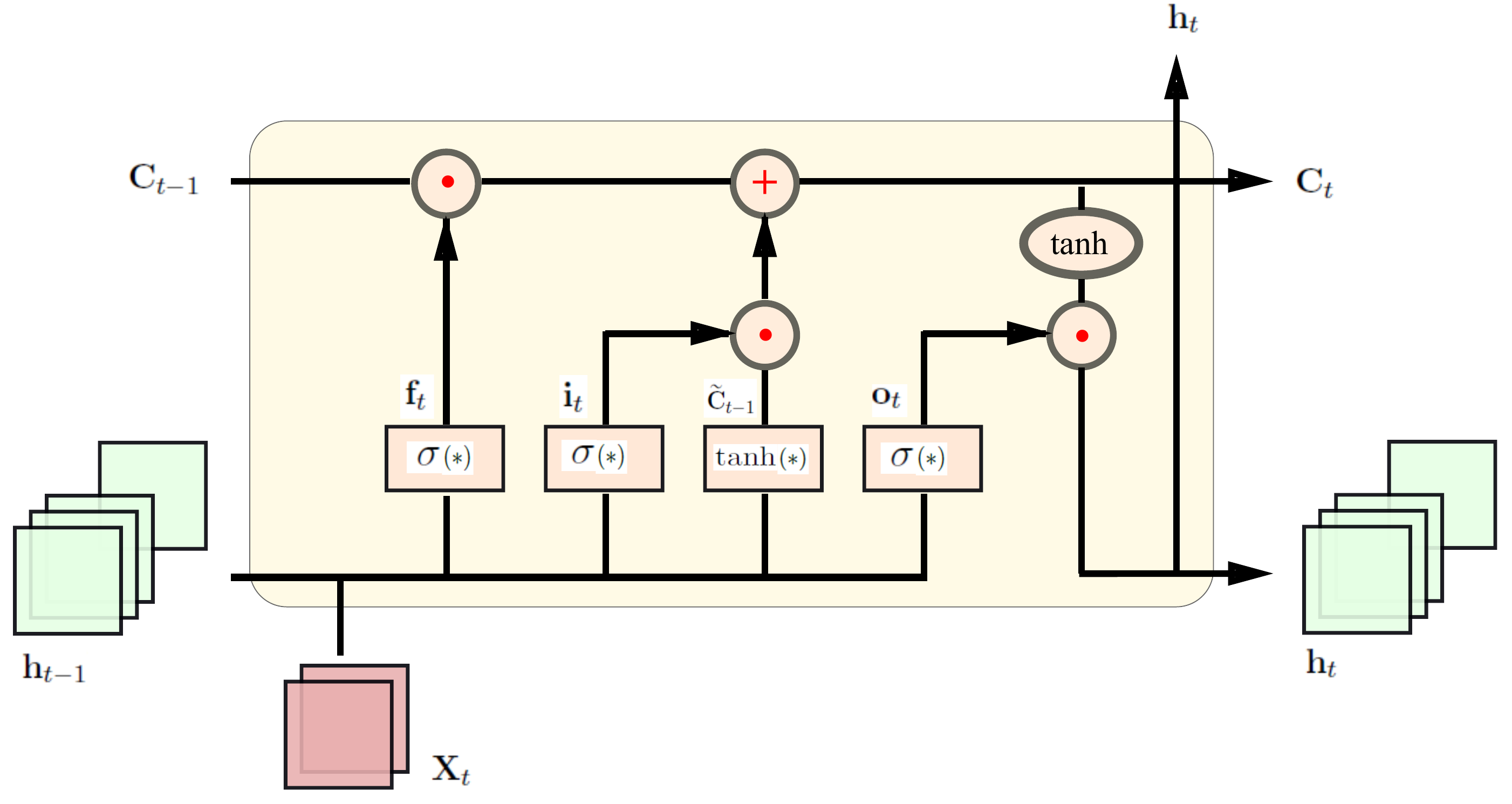}
      \caption{\label{fig:ConvLSTM} The single ConvLSTM cell at time $\mathrm{t}$}
    \end{center}\hspace{2pc}%
\end{figure}

Long short-term memory (LSTM) \cite{Hochreiter} is a class of Recurrent Neural networks (RNNs) \cite{Dupond}. The gradient vanishing problem for long-term temporal dependencies in RNNs is solved in LSTMs. LSTMs have a memory cell that can maintain information in memory for a long period of time and also gates that allow for better control over the gradient flow by forgetting, updating, and outputting part of the needed information. These gates, in fact, enable better preservation of long-term time dependencies. To consider a simultaneous spatial and temporal learning framework, an extension of LSTMs named ConvLSTMs \cite{Shi} is employed, which is basically the Convolutional Neural Networks (CNNs) extension of LSTMs. Therefore, convolutional layers are employed instead of the fully-connected NNs (dense layers) in gated operations because of their better representational capability of spatial connections. Thus, the applied ConvLSTM extended form of the long
short-term memory (LSTM) is a spatio-temporal that is developed for the purpose of sequence-to-sequence learnings. Figure \ref{fig:ConvLSTM} demonstrates a typical graphic of ConvLSTMs. \\

In Figure \ref{fig:ConvLSTM}, $\tnb{X}_{\mathrm{t}}$ stands for the input tensor. The hidden state and cell state are indicated with $\tnb{h}_{\mathrm{t}}$ and $\tnb{C}_{\mathrm{t}}$, respectively, to be updated at time $\mathrm{t}$. 
The ConvLSTM cell consists of four gate variables in input-to-state transition and state-to-state transition. The forget gate and input gate are indicated with $\{\tnb{f}_{\mathrm{t}}\}$ and $\{\tnb{i}_{\mathrm{t}}\}$, respectively, at time $\mathrm{t}$. The other two gates, an internal cell and an output gate, are also denoted with $\{\tilde{\tnb{C}}_{\mathrm{t}}\}$ and $\{\tnb{o}_{\mathrm{t}}\}$, respectively. 
\\

Because the sigmoid activation function $\sigma(\cdot)$ is used for the gates, the mapping outputs to values between $0$ and $1$. Therefore, the forget gate layer adaptively clears the memory information in the cell state $\{\tnb{C}_{\mathrm{t-1}}\}$. The memory stored in
cell state originates from the cooperation between the input gate layer and the internal cell state,
where the internal cell state is a new cell candidate created from the hyperbolic tangent activation
layer (i.e., $\ops{tanh}(\cdot)$) and the input gate layer decides the information propagating into the cell state. Lastly, the output gate layer filters and regulates the cell state for the final output variable/hidden
state. The updating ConvLSTM is governed by the mathematical formulations which are described in Equation \ref{eq:ConvLSTM}.\\

\begin{subequations}
\label{eq:ConvLSTM}
\begin{align}
& \tnb{i}_{\mathrm{t}} = \sigma (\tnb{W}_{\mathrm{i}} * [ \tnb{X}_{\mathrm{t}}, \tnb{h}_{\mathrm{t-1}} ] + \vek{b}_{\mathrm{i}}  )
\\ 
& \tnb{f}_{\mathrm{t}} = \sigma (\tnb{W}_{\mathrm{f}} * [ \tnb{X}_{\mathrm{t}}, \tnb{h}_{\mathrm{t-1}} ] + \vek{b}_{\mathrm{f}}  )
\\
& \tilde{\tnb{C}}_{\mathrm{t-1}} = \ops{tanh} (\tnb{W}_{\mathrm{c}} * [ \tnb{X}_{\mathrm{t}}, \tnb{h}_{\mathrm{t-1}} ] + \vek{b}_{\mathrm{c}}  )
\\ 
& \tnb{C}_{\mathrm{t-1}} = \tnb{f}_{\mathrm{t}} \odot \tnb{C}_{\mathrm{t-1}} + \tnb{i}_{\mathrm{t}} \odot \tilde{\tnb{C}}_{\mathrm{t-1}}
\\ 
& \tnb{o}_{\mathrm{t}} = \sigma (\tnb{W}_{\mathrm{o}} * [ \tnb{X}_{\mathrm{t}}, \tnb{h}_{\mathrm{t-1}} ] + \vek{b}_{\mathrm{o}}  )
\\ 
& \tnb{h}_{\mathrm{t}} = \tnb{o}_{\mathrm{t}} \odot \ops{tanh}( \tnb{C}_{\mathrm{t}} )
\end{align}
\end{subequations}

In Equation \ref{eq:ConvLSTM}, $*$ indicates the convolutional operation and $\odot$ denotes the Hadamard product. Also, $\{\tnb{W}_{\mathrm{i}}, \tnb{W}_{\mathrm{f}}, \tnb{W}_{\mathrm{c}}, \tnb{W}_{\mathrm{o}}\}$ are the weight parameters of the model for the corresponding filters where $\{\vek{b}_{\mathrm{i}}, \vek{b}_{\mathrm{f}}, \vek{b}_{\mathrm{c}}, \vek{b}_{\mathrm{o}}\}$ represent bias vectors.\\

\subsection{Additional Techniques}
\label{subsection:Additional Techniques}

The input and label data for a typical storm were shown earlier in Section \ref{section:Storm Surge Prediction Problem Characteristics} in Figure \ref{fig:inputs0} and Figure \ref{fig:output200}. 
As discussed earlier, Figure \ref{fig:inputs0} clearly shows that some of the key input data (size and intensity of storm) do not vary substantially before the storms makes landfall, creating significant challenges for effective training of the model using them. A typical ConvLSTM model is not able to learn from the data as it carries out one-to-one learning \cite{Sutskever}. Initial attempts to train a ConvLSTM model to predict the upcoming storm surge based on the desired inputs were unsuccessful because a sequence to sequence prediction model \cite{Sutskever} was required for the studied datasets. Therefore, a few techniques were developed to adapt the model to the datasets, so the model can be trained effectively. 
In the following, we discuss these techniques.\\

To accommodate the requirement to establish a sequence-to-sequence prediction model \cite{Sutskever}, 
an encoder-decoder, a popular approach  
of organizing recurrent neural networks for sequence-to-sequence prediction applications, is used. Encoder-decoder models are very capable with the sequential data since the LSTM layer is developed to work with sequential model \cite{Graves, Sutskever}. With a finely tuned LSTM layer, we can make a whole network perform appropriately with the sequential information of the data by making the network memorize the sequence. The Encoder-decoder modeling involves two recurrent neural networks, one to encode the source sequence: one for reading the input sequence, called the encoder and a second to decode the encoded source sequence into the target sequence, decoding the fixed-length vector and outputting the predicted sequence, called the decoder. Here our original model is combined with the encoder-decoder network model to build a high-performance model for the desired sequential data.\\

Inspired by the forward Euler scheme, a global residual connection is also designed. The residual connection is between the
input state variable $\tnb{u}_{\mathrm{i}}$ and the output variable $\tnb{u}_{\mathrm{i+1}}$. The learning process at time instant $\mathrm{t}_{\mathrm{i}}$ is formulated as $\tnb{u}_{\mathrm{i+1}}$ = $\tnb{u}_{\mathrm{i}} + \delta \mathrm{t} \cdot \tnb{\mathcal{NN}} [\tnb{u}; \tnb{\theta}  ]  $, where $\tnb{\mathcal{NN}}$ denotes the trained network operator and
$\mathrm{t}$ is the time interval. Based on this formulation, the output state variable $\tnb{u}_{\mathrm{i+1}}$ at time instant $\mathrm{t}_{\mathrm{i}}$ switches into the input variable at $\mathrm{t}_{\mathrm{i+1}}$. These residual connection networks is the second technique employed to improve the performance of the developed model.\\ 

\begin{figure}[H]
    \begin{center}{}
      \includegraphics[width=4.25in]{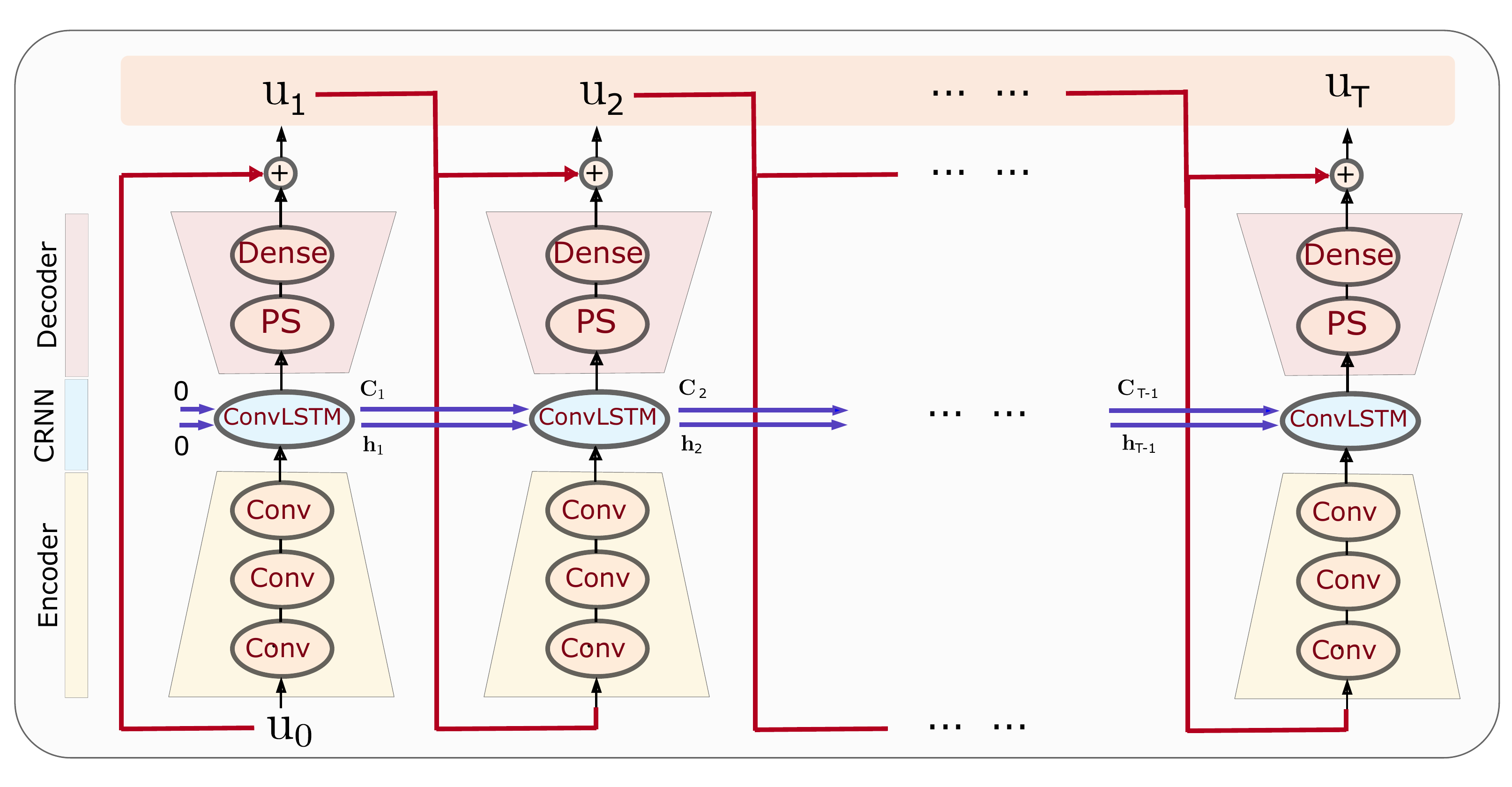}
      \caption{\label{fig:ConvLSTMstructure} The developed Convolutional Recurrent Neural Network (CRNN) structure}
    \end{center}\hspace{2pc}%
\end{figure}

The other technique we leverage is pixel shuffle \cite{Shi1}, which is an upsampling strategy. Pixel shuffle maintains satisfactory reconstruction accuracy in image and video super-resolution tasks without high computational and memory costs \cite{Shi1}. In comparison to deconvolution \cite{Shan} which always needs more layers to reach the expected resolution, pixel shuffle has lower computational complexity. Beyond that, another advantage of pixel shuffle is that it introduces fewer checkerboard artifacts compared with deconvolution \cite{Odena}. 
The final developed model structure, incorporating all aforementioned advances, is shown in Figure \ref{fig:ConvLSTMstructure} where PS stands for pixel shuffle.

\subsection{Training Process}
\label{subsection:Training Process}

Considering the special shape of inputs shown for one storm in Figure \ref{fig:inputs0}, i.e. the fact that, as stressed earlier, certain inputs are not changing very much over time, special attention needs to be given to data standardization. We first standardize these four inputs separately, one by one, using $\mathrm{x}^{\prime} = \frac{\mathrm{x} - \mu}{\sigma}$ where $\mu$ and $\sigma$ are the mean and standard deviation, respectively. The label data (the outputs), shown in Figure \ref{fig:output200}, are also normalized in a way that they centralized around zero by means of the hyperbolic function. The model experiences a faster convergence for such normalized data.\\

Before discovering the data through the ConvLSTM cell, we pass the input data through the encoder to study the entire sequence of the data. The encoder contains three convolutional layers where the ReLU activation function is employed for these layers. The kernel size, padding size, and stride size for these three layers are $4\times 4$,  $1\times 1$, and  $2\times 2$, respectively. These three layers' input and output channels are receptively 2 and 16, 16 and 32, and 32 and 64. Right before the encoder with these three convolutional layers, three linear layers are designed, fed with four inputs, and outputted the same dimension of label data, 4800 elements. For these linear layers, the hyperbolic activation function, $\ops{tanh}$, is used. Once the data are provided in the latent space, a ConvLSTM cell is employed where the kernel size, padding size, and stride size are $5\times 5$, $2\times 2$, and $1\times 1$, respectively. The input and output channels for the ConvLSTM cell are both 64. Note that model training in the latent space, where the ConvLSTM layer is the optimal space to train the model. The only two layers in the decoder are upsampling through pixel shuffle explained in Section \ref{subsection:Additional Techniques} and a final linear layer that outputs the same shape of data as label data. It should be pointed out that the pixel shuffling decreases the channel size from 64 to 1 as a pixel shuffle layer with an upscale factor 8 is applied. It also increases the height and width of data by 8. Table \ref{tab:layers} shows all the employed layers, filter sizes, and outputs of each layer separately. The training is carried out with a batch size of 100. Therefore, each batch contains 100 storm data, and the steps above are repeated for all 125 time steps.\\

\begin{table}[H]
\caption{Convolutional recurrent neural network model architecture}
\vspace{-0.5cm}
\begin{center}
\scalebox{0.9}{
\begin{tabular}{l*{3}{c}r }
\hline
Cell      & Layer & Filter/Upscale factor 
 & Output &\\
\hline
  $\text{Input} $  & $\text{}$& $ $ & $\text{[100, 1, 4]}$ &\\
  $ $                       & $\text{Dense}$& $ $ & $\text{[100, 1, 40]}$ &\\
  $ $                       & $\text{Dense}$& $ $ & $\text{[100, 1, 400]}$ &\\
  $ $                       & $\text{Dense}$& $ $ & $\text{[100, 1, 4800]}$ &\\
  $ $                       & $\text{Reshape}$& $ $ & $\text{[100, 1, 120, 40]}$ &\\
  $ \text{Encoder}$  & $\text{Convolutional}$& $\text{[4, 4, 16]} $ & $\text{[100, 16, 60, 20]}$ &\\
    $ $                       & $\text{Convolutional}$&  $\text{[4, 4, 32]} $ & $\text{[100, 32, 30, 10]}$ &\\
    $ $                       & $\text{Convolutional}$& $\text{[4, 4, 64]} $ & $\text{[100, 64, 15, 5]}$ &\\
    $\text{ConvLSTM}$                        & $\text{ConvLSTM}$& $\text{[5, 5, 64]} $ & $\text{[100, 64, 15, 5]}$ &\\
    $\text{Decoder}$                        & $\text{Pixel Shuffle}$& $\text{[8]} $ & $\text{[100, 1, 120, 40]}$ &\\
  $ $                       & $\text{Reshape}$& $ $ & $\text{[100, 1, 4800]}$ &\\
    $\text{}$                        & $\text{Dense}$& $ $ & $\text{[100, 1, 4800]}$ &\\   
    $\text{Output}$                        & $\text{}$& $ $ & $\text{[100, 1, 4800]}$ &\\   
\hline
\end{tabular}
}
\label{tab:layers}
\end{center}
\end{table}

The model is trained with $35000$ epochs, and the learning rate is selected as $1\mathrm{e}-4$. The $\mathrm{L}_2$ norm loss function is minimized over the epochs by the mini-batch gradient descent method as follows.

\begin{subequations}
\label{eq:L2}
\begin{align}
& \mathcal{L} (\tnb{W}, \vek{b}) = || {\mathcal{C} (t, \vek{x}, \tnb{\theta}) - \vek{z}_L (t, \vek{x}, \tnb{\theta} ; \tnb{W}, \vek{b})  } ||_{\mathrm{L}_2(\Omega)}
\\ 
& \tnb{W}^*, \vek{b}^* = \underset{\tnb{W}, \vek{b}}{\text{arg}\min}  \,\, \mathcal{L} (\tnb{W}, \vek{b})
\end{align}
\end{subequations}

In Equation \ref{eq:L2}, $\mathcal{L}(\cdot)$ stands for the loss function and the $\mathrm{L}_2$ norm is indicated with $|| \cdot||_{\mathrm{L}_2(\Omega)}$. The CFD solution (storm surge database predictions) is denoted with $\mathcal{C}(t, \vek{x}, \tnb{\theta})$ and $\tnb{W}^*, \vek{b}^*$ represent the (sub)optimal neural network parameters, the weights and biases obtained from the optimization problem.\\

\begin{figure}[H]
    \begin{center}{}
      \includegraphics[width=3.0in]{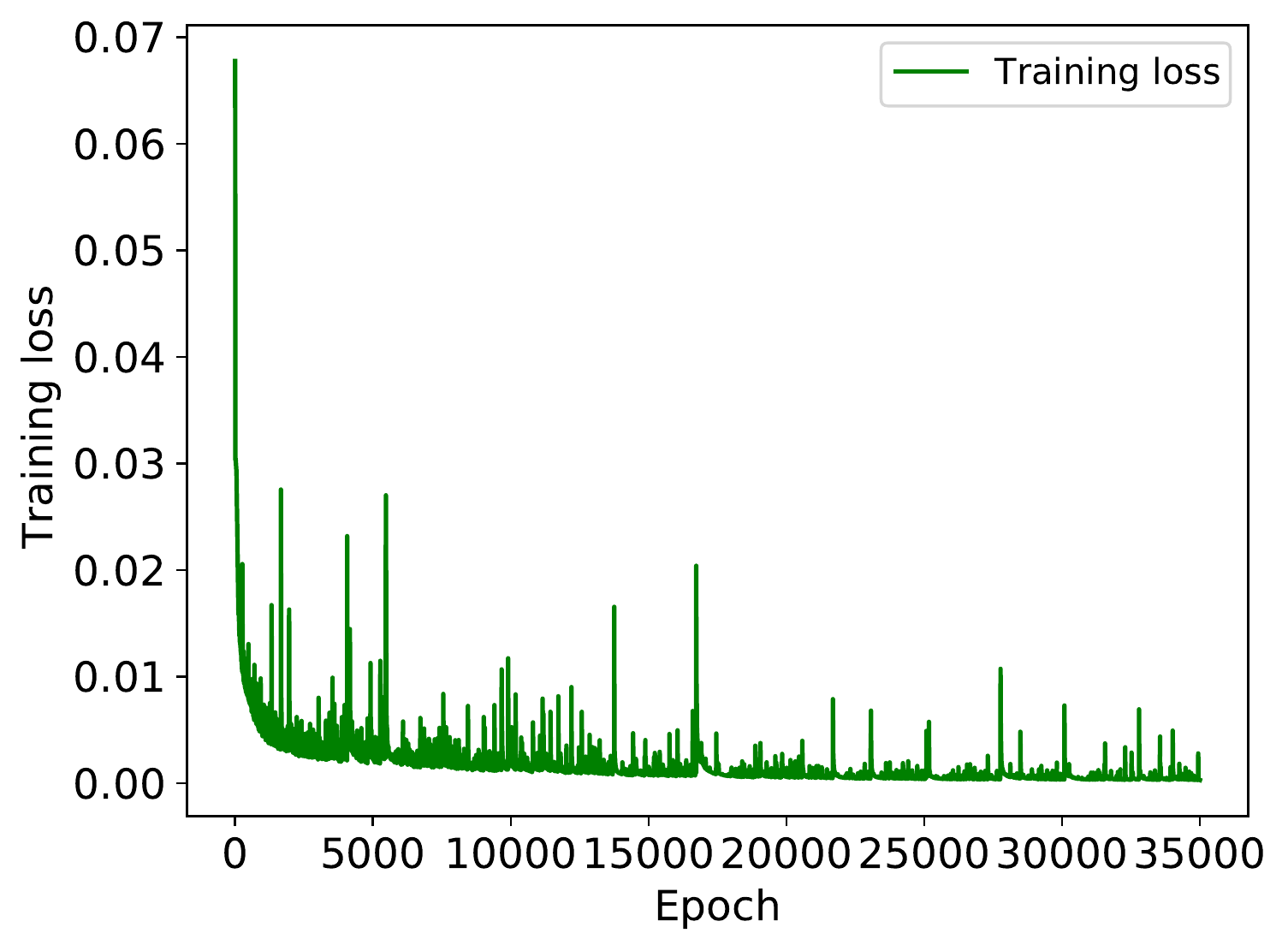}
      \caption{\label{fig:Loss} Loss function over epochs}
    \end{center}\hspace{2pc}%
\end{figure}

The hyper-parameters are set initially randomly. Many models with different sets of hyper-parameters are run in parallel to find the models whose loss values converge over epochs. Once the trainable models with specified hyper-parameters are determined, the hyper-parameters are evaluated in random search \cite{Bergstra} to find the optimal set of hyper-parameters. The loss function over epoch numbers for the optimal model trained on the studied storm datasets is shown in Figure \ref{fig:Loss}.\\

As Figure \ref{fig:Loss} shows, the loss error decreases continuously over epochs from about $7\mathrm{e}-2$ to $3\mathrm{e}-4$, which shows a significant reduction. Once the model is trained, it is ready to predict the storm surges for new storms with the provided input values. The next section assesses the train model evaluation and the predicted storm surge's accuracy.

\section{Model Evaluation}
\label{section:Model Evaluation}
%
%
%
%
%
%
%
%
%
%
Eight synthetic storms within the original database, not utilized in the training phase, are now used to validate the performance of the developed convolutional recurrent neural network model. The predicted surges are compared to the label test data, corresponding to the simulated surge for the same SPs and time steps utilized in the model development. An alternative surrogate model implementation is also considered in this section, a Gaussian Process emulator that has been previously developed. 
Approach utilizes a simplified parameterization of the storm input, using instantaneous storm features close to landfall to characterize each storm, and considers independent predictions for the surge for each SP or time-step, using principal component analysis to incorporate spatio-temporal correlation features in the surge predictions. Further details for this formulation are discussed in \cite{Jia1}. The root square mean errors (RMSE) of the test set is reported in Table \ref{tab:Test}, separately for each storm.

\begin{table}[H]
\caption{RMSE of test datasets}
\vspace{-0.5cm}
\begin{center}
\scalebox{0.6}{
\begin{tabular}{l*{8}{c}r}
\hline
  \text{} & \text{Test 1}  &  \text{Test 2} & \text{Test 3} & \text{Test 4} &\text{Test 5} & \text{Test 6} & \text{Test 7} &\text{Test 8} &\\
\hline
   $\text{CRNN}$ & $1.089\mathrm{e}-2$ & $2.799\mathrm{e}-2$ & $6.914\mathrm{e}-2$ & $2.749\mathrm{e}-2$ & $4.351\mathrm{e}-2$ & $6.677\mathrm{e}-2$ & $4.946\mathrm{e}-2$ & $1.302\mathrm{e}-1$ &\\ 
   $\text{GP}$ & $2.111\mathrm{e}-2$ & $8.912\mathrm{e}-2$ & $1.339\mathrm{e}-1$ & $3.743\mathrm{e}-1$ & $5.178\mathrm{e}-2$ & $9.386\mathrm{e}-2$ & $4.663\mathrm{e}-2$ & $2.177\mathrm{e}-1$ &\\  
\hline
\end{tabular}
}
\label{tab:Test}
\end{center}
\end{table}

As it is shown in Table \ref{tab:Test} by comparing the RMSE of the predictions by the developed neural network model and the Gaussian Process for every single storm, it can be inferred that the neural network model offers greater accuracy storm surge predictions than the Gaussian process for all of the storm datasets. The average RMSE of the predictions for these eight tests is $5.312\mathrm{e}-2$ and for Gaussian process is $1.285\mathrm{e}-1$, which shows at least a two-times less error in total.\\

\begin{figure}[H]
    \begin{center}{}
      \includegraphics[width=2.5in]{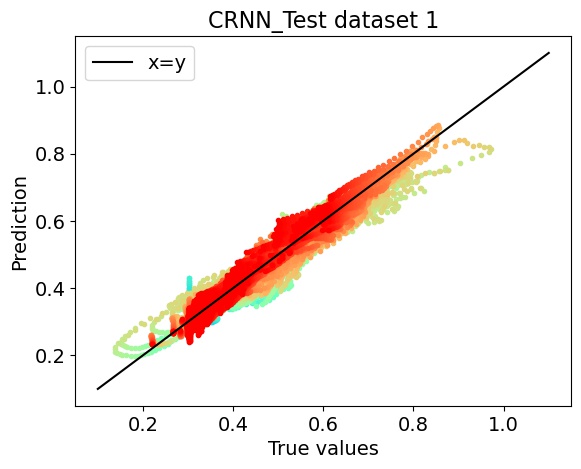}  
      \includegraphics[width=2.5in]{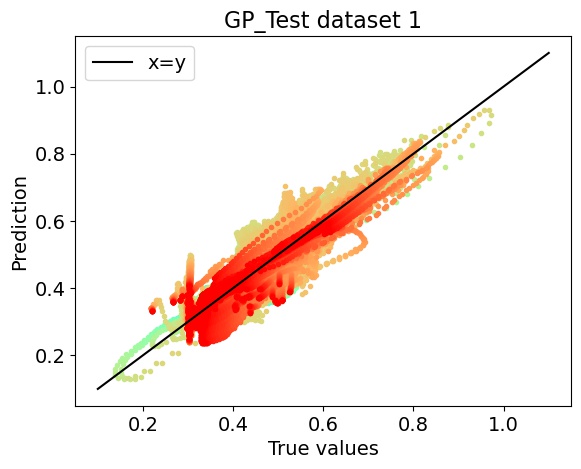}      
      \caption{\label{fig:predtrue} Storm surge prediction vs. true test data provided by CRNN and GP}
    \end{center}\hspace{2pc}%
\end{figure}

For one of the test datasets, we look at true values and predictions provided by the developed neural network model and the Gaussian process method in Figure \ref{fig:predtrue}. In this Figure, the line $\mathrm{x=y}$ is also plotted to better present the correlation of predicted and true values.\\

As it can be seen in Figure \ref{fig:predtrue}, the predicted surge values by the neural network model over true values are more concentrated around the line $x=y$ than the predicted values by the Gaussian process. In other words, the predicted surge values by the Gaussian process versus the true surge values are more scattered than the predicted values by the developed neural network model over their corresponding true surge values. It means the predictions by the developed model could estimate the true values more accurately than the predictions by the Gaussian process. \\

\begin{figure}[H]
    \begin{center}{}
      \includegraphics[width=2.5in]{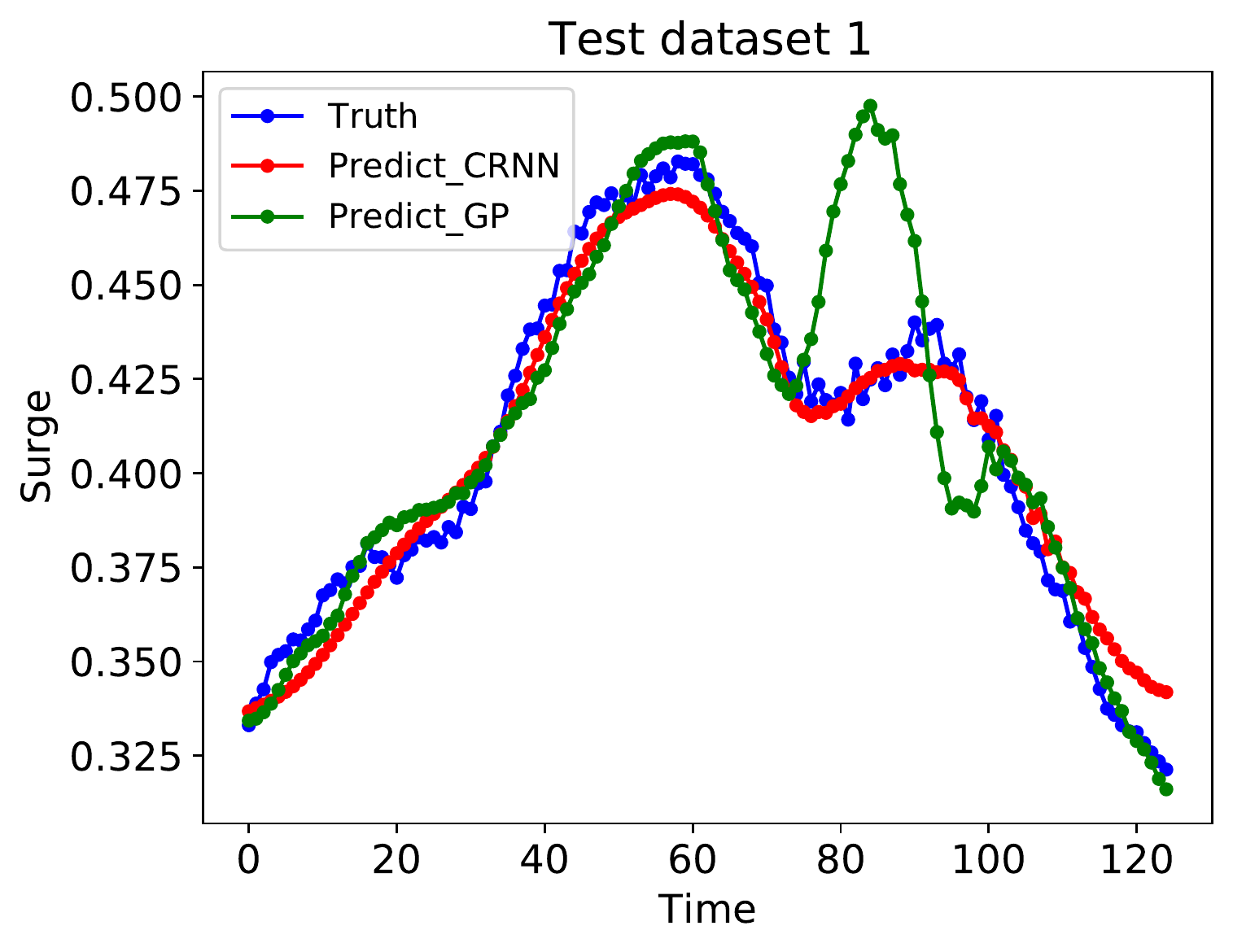}  
      \includegraphics[width=2.5in]{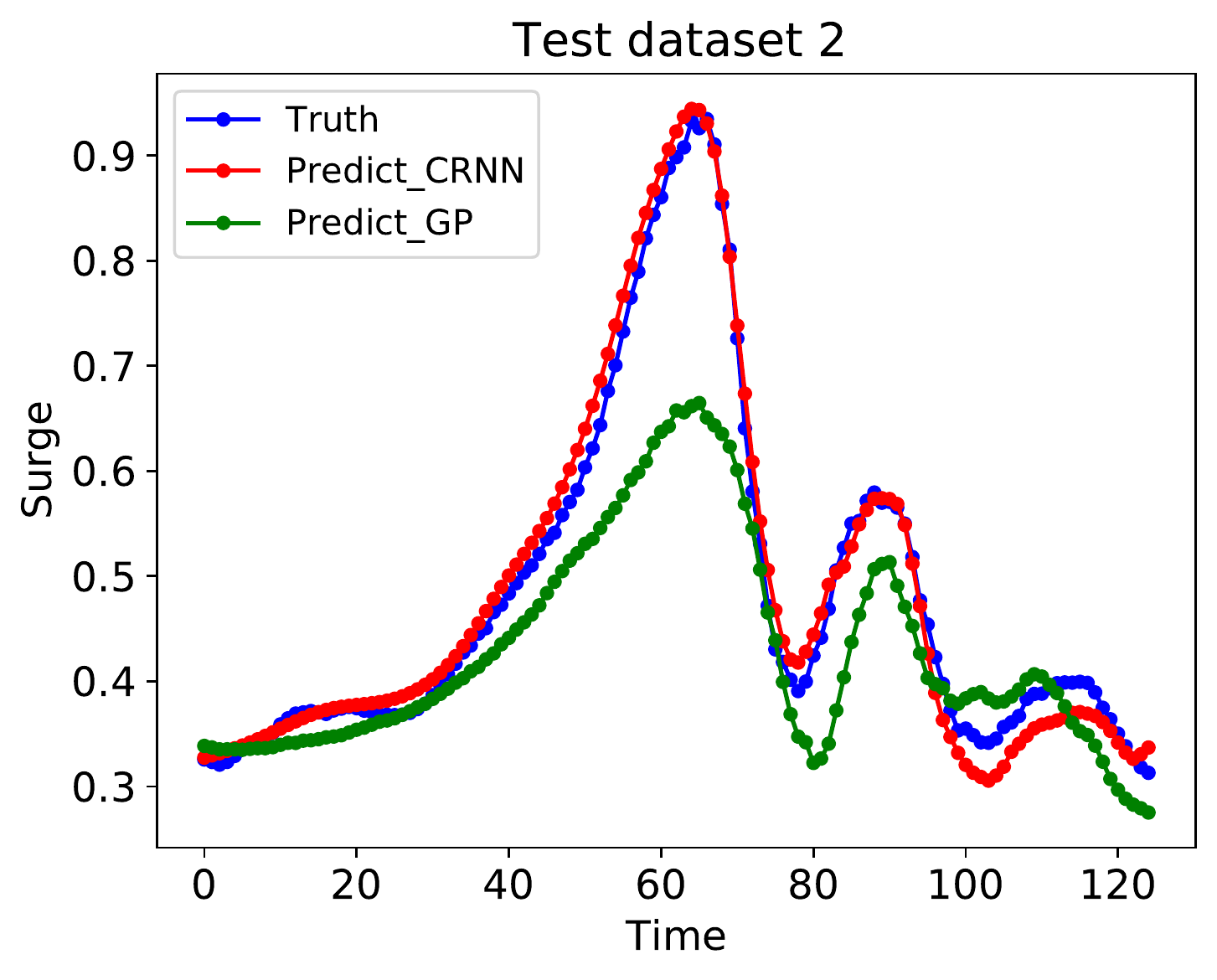} 
      \includegraphics[width=2.5in]{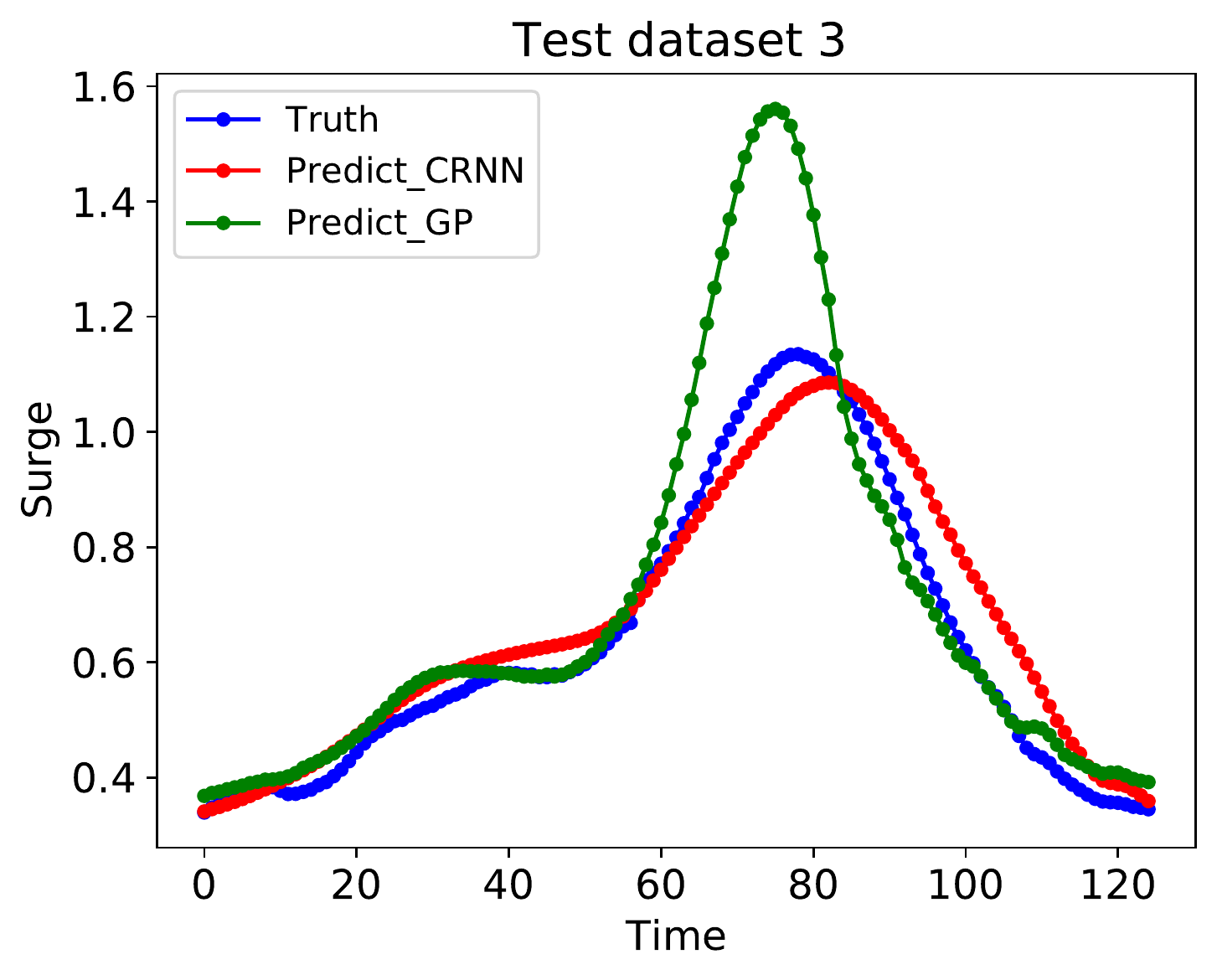} 
      \includegraphics[width=2.5in]{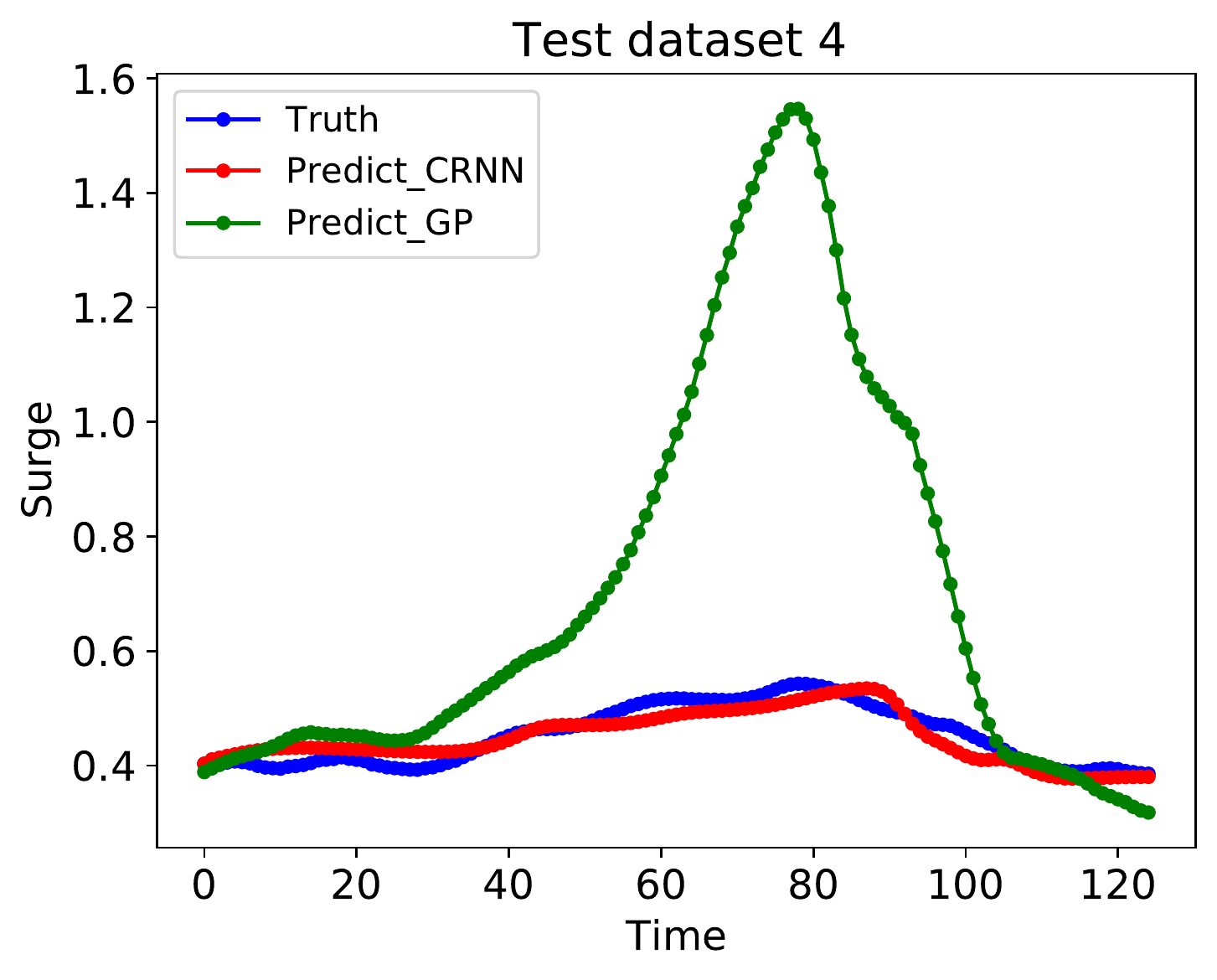} 
      \includegraphics[width=2.5in]{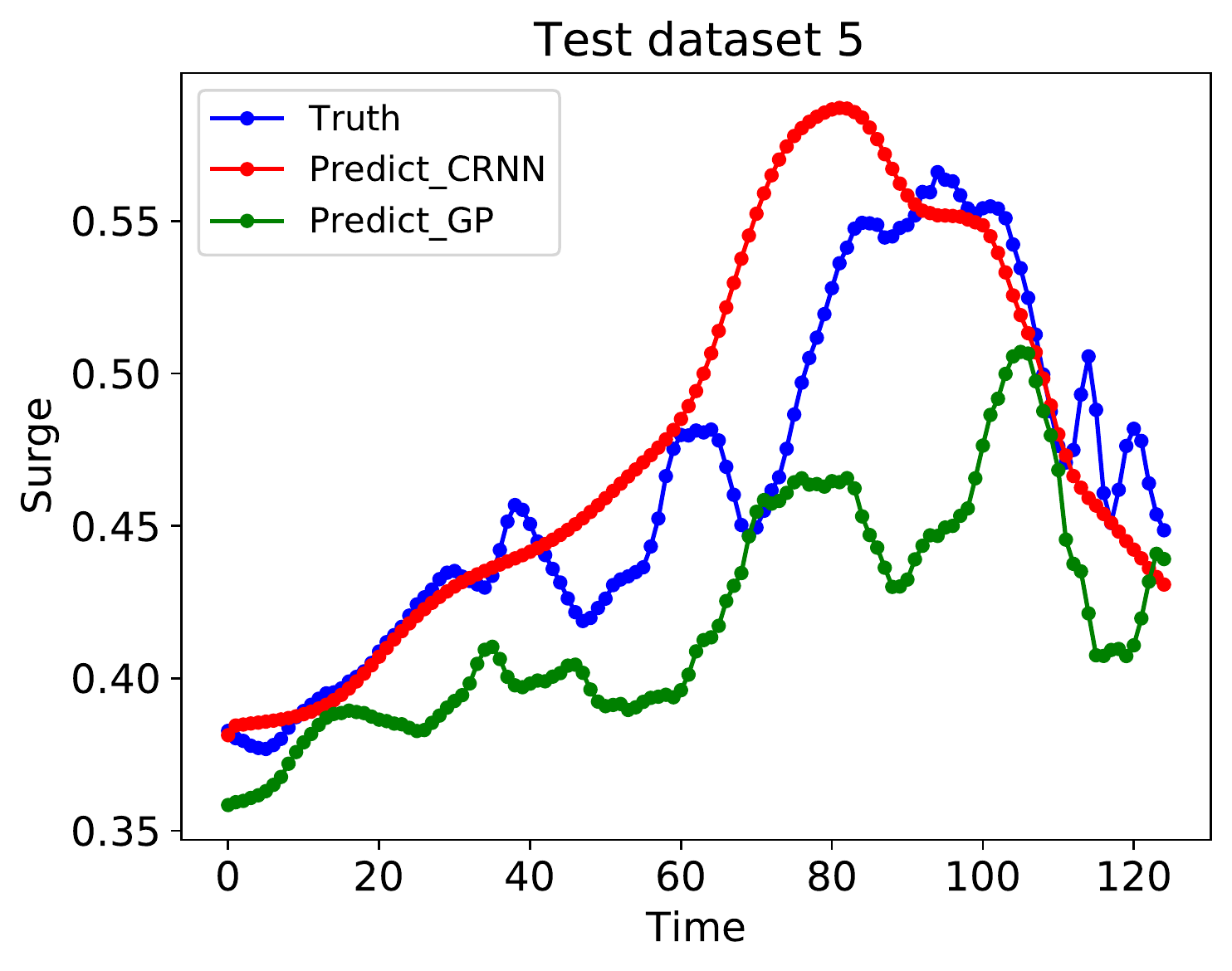} 
      \includegraphics[width=2.5in]{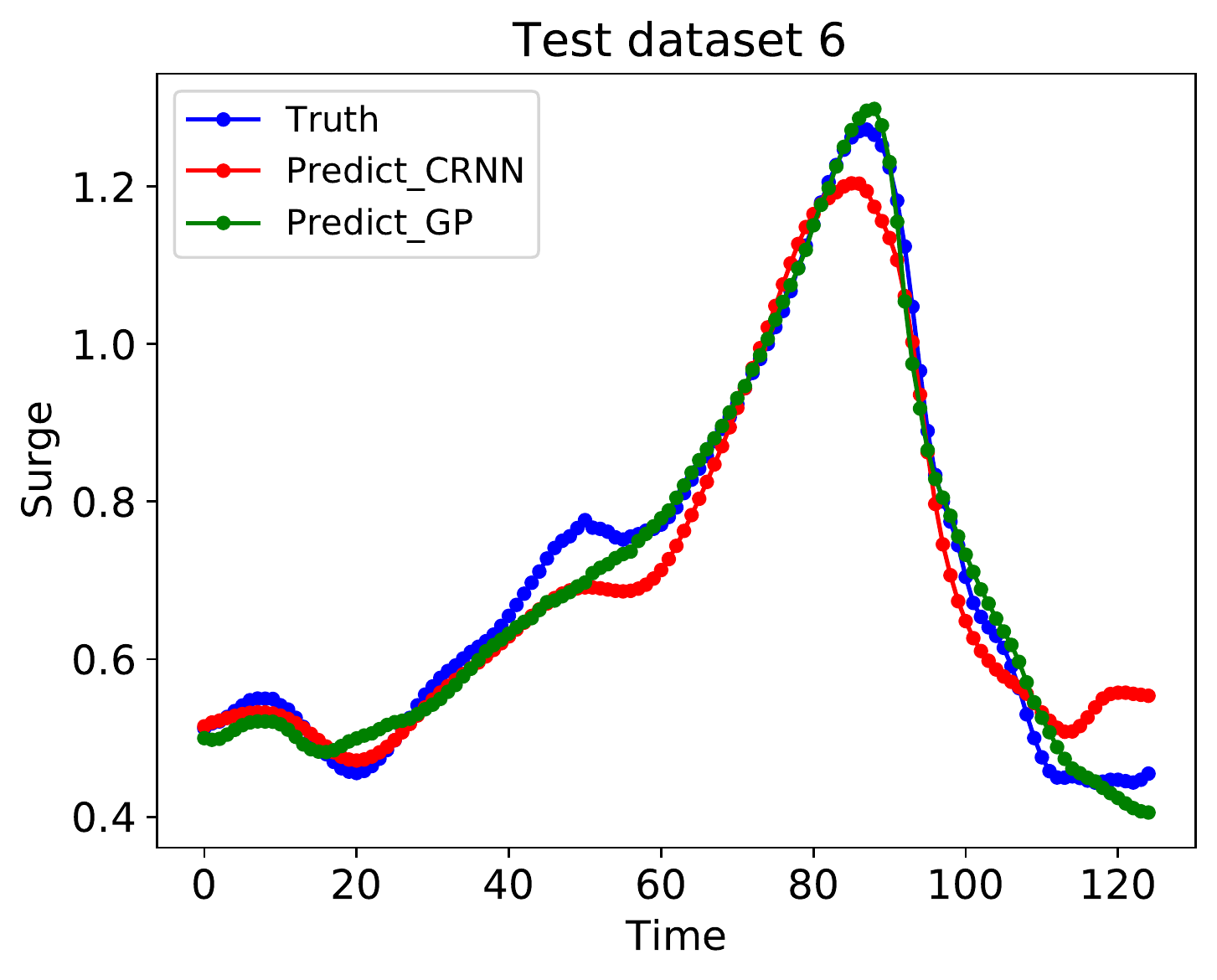} 
      \includegraphics[width=2.5in]{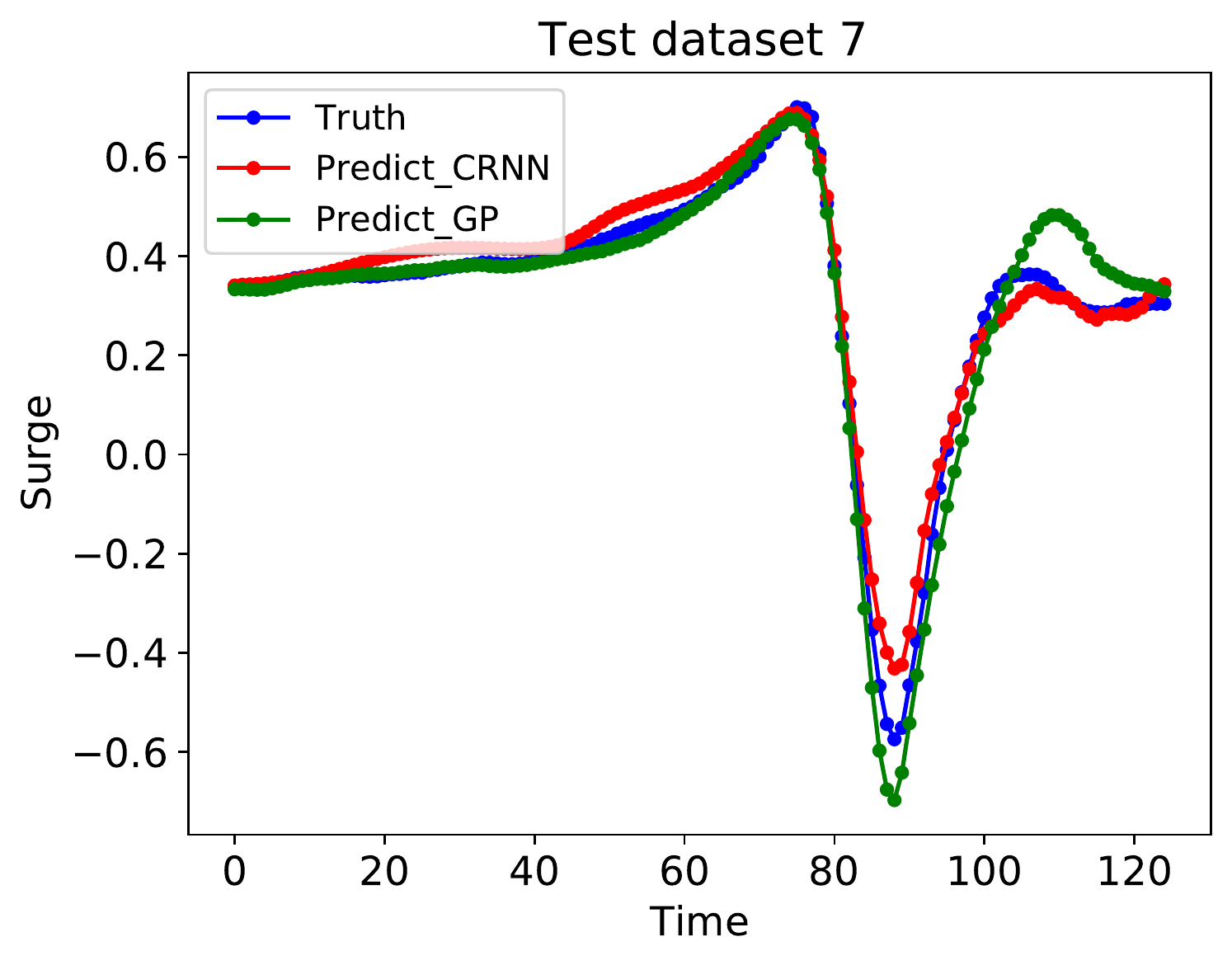} 
      \includegraphics[width=2.5in]{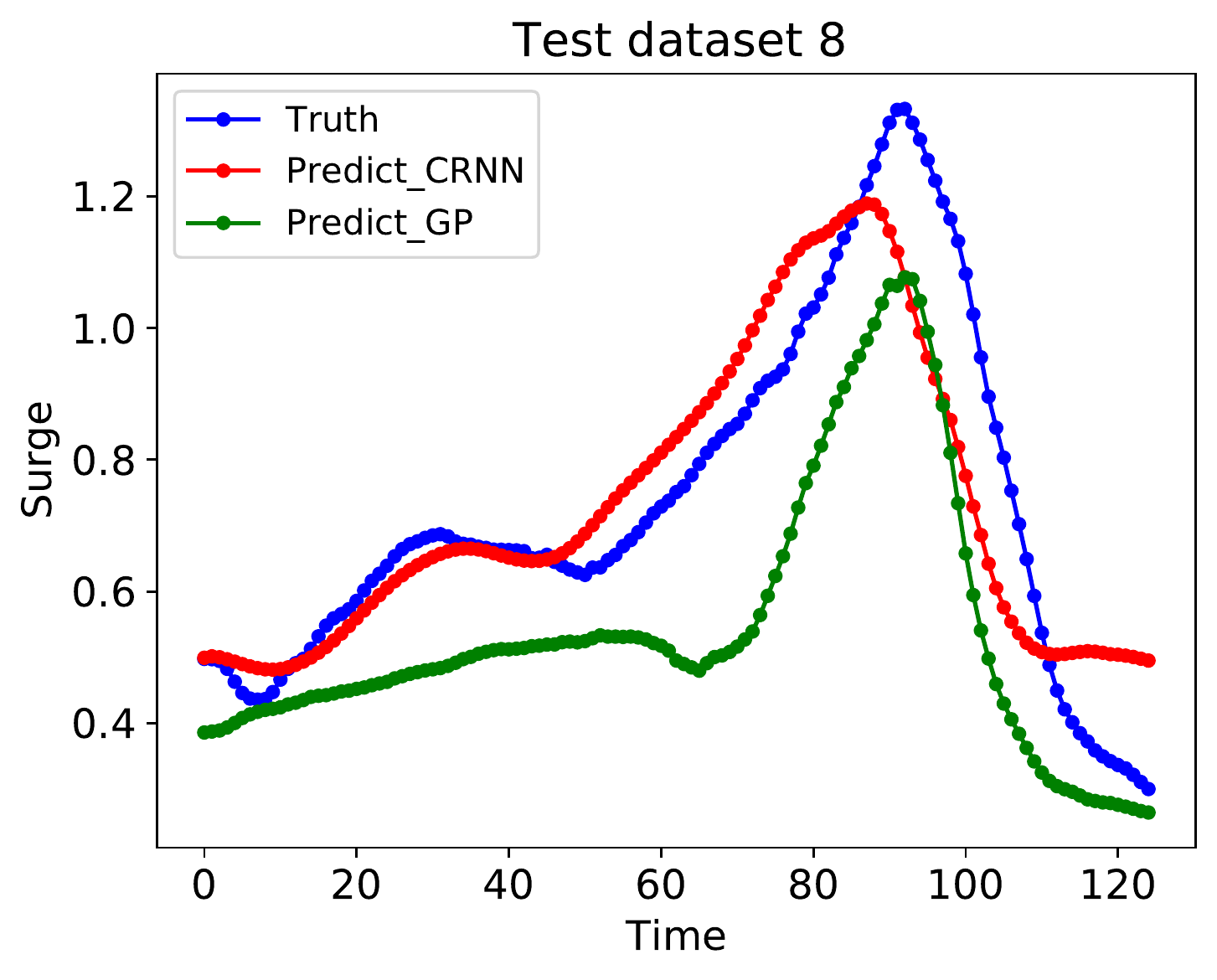}       
      \caption{\label{fig:CRNN_GP_middle} Storm surge predictions for one grid SP in different test storms for a grid point in the coast middle layers}
    \end{center}\hspace{2pc}%
\end{figure}

Moreover, to further confirm our observations, we compare the predictions by the developed neural network model and the Gaussian process with the true storm surges of the eight test datasets for the same station (grid) in the coast in all the test datasets in Figure \ref{fig:CRNN_GP_middle}. Note that this grid is chosen from the middle layers of the region's grids on the coast.\\

As it is seen in Figure \ref{fig:CRNN_GP_middle}, the storm surge predictions computed by the developed convolutional recurrent neural network in all the studied tests are very close to the true values for the entire time interval. 
The Gaussian process provides less accurate storm surge predictions as the predictions are generally far away from the true surge values. However, the Gaussian process has partially learned the data trend, and its surge predictions can somewhat mimic the surge true values' trend. 
Two more SPs from very early layers of the coast and end layers of the coast are chosen, and the predictions provided by these two approaches are compared with the true values in Appendix \ref{Test Grids}. 
By comparing the surge predictions by the two studied methods and the true surge values, the observation mentioned above can be generalized for all the grids on the coast.

\section {Conclusions}
\label{section:Conclusions}

This study examined the development of a neural network for emulating time-series surge predictions using a database of synthetic storm simulations. The developed convolutional recurrent neural network model is enriched by an encoder-decoder model, so that the developed model takes the entire sequence of the data into account. Therefore, the entire storm surge can be predicted based on the storm-driven parameters' complete history. The encoder-decoder add-on ultimately makes the developed neural network model a sequence to sequence (seq2seq) storm surge forecast model. Also, the model's performance is increased by incorporating a residual connection network. Several techniques are also applied in the training process to improve performance.
Overall, the spatial and temporal correlations of the data are captured by employing convolutional neural network layers and the recurrent neural network, respectively, through a ConvLSTM cell. The ConvLSTM cell is trained on the data provided in the latent space right between encoder and decoder cells, accommodating better learning for the ConvLSTM cell.
In contrast to previous storm surge prediction studies where machine learning methods were predominantly used as black boxes and surge for a few representative stations was only predicted, the aforementioned formulation allows us to predict surge for all the save points within the domain of interest by establishing problem-specific advances for the neural network implementation. 
Furthermore, through these formulations, the correlations of data both spatially and temporally are learned by the model to enhance prediciton accuracy, something that again contrasts to previous studies. 
The evaluation of the trained model on test datasets show that the model can accurately predict the storm surge. The develop model can, ultimately, accommodate fast predictions for the time-series surge evolution, driven by track/size/intensity storm input features, and can be used to support efficient risk assessment and emergency response management operations.   \\

\noindent \textbf{\textit{Acknowledgement}} \\
Authors would like to thank the Army Corp of Engineers, Coastal Hydraulics Laboratory of the Engineering Research and Development Center for providing access to the storm surge data, though the coastal hazards system (https://chs.erdc.dren.mil/), that were used in the illustrative case study. 

\newpage



\begin{appendices}

\section{Test Grids}
\label{Test Grids}

\begin{figure}[H]
    \begin{center}{}
      \includegraphics[width=2.5in]{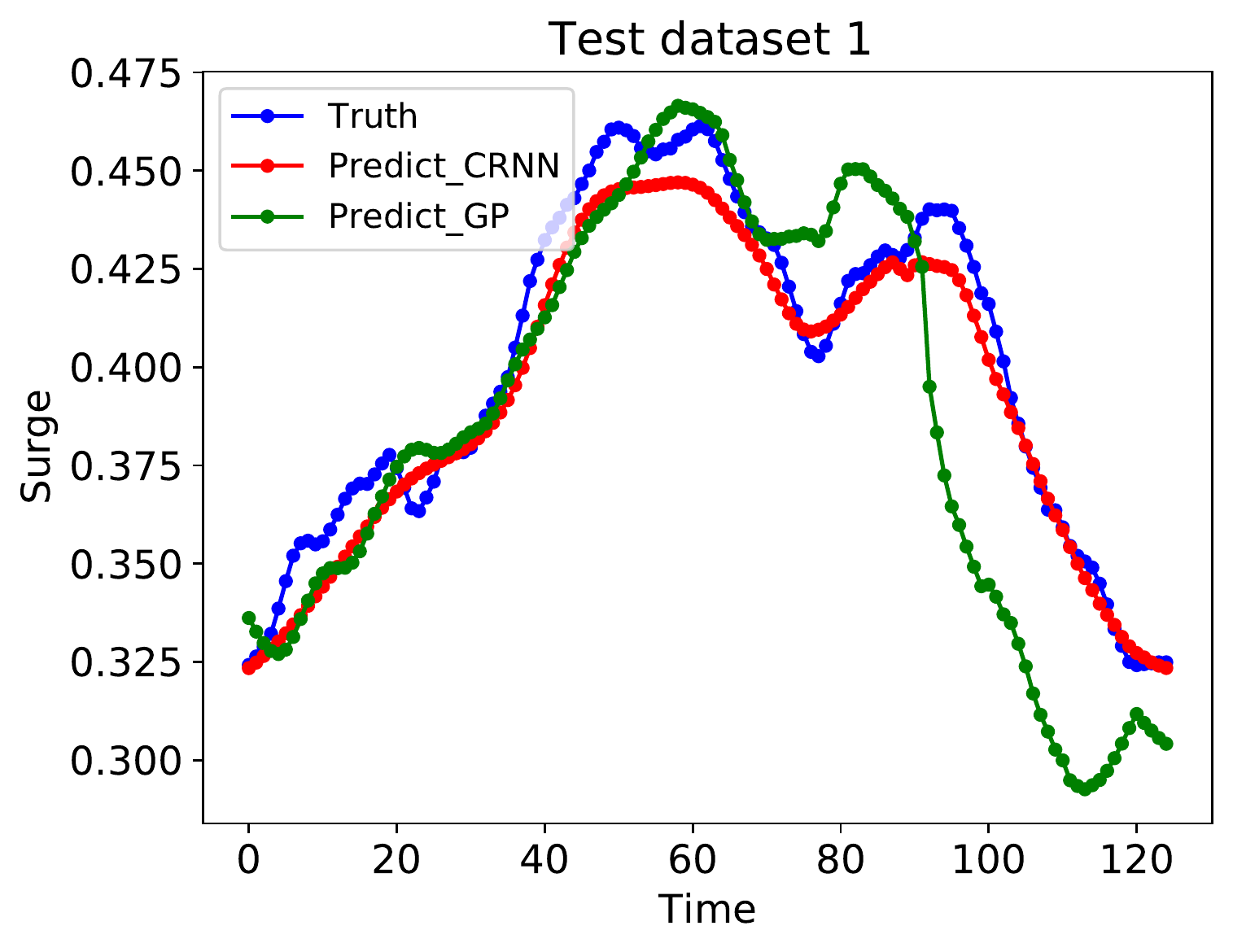}  
      \includegraphics[width=2.5in]{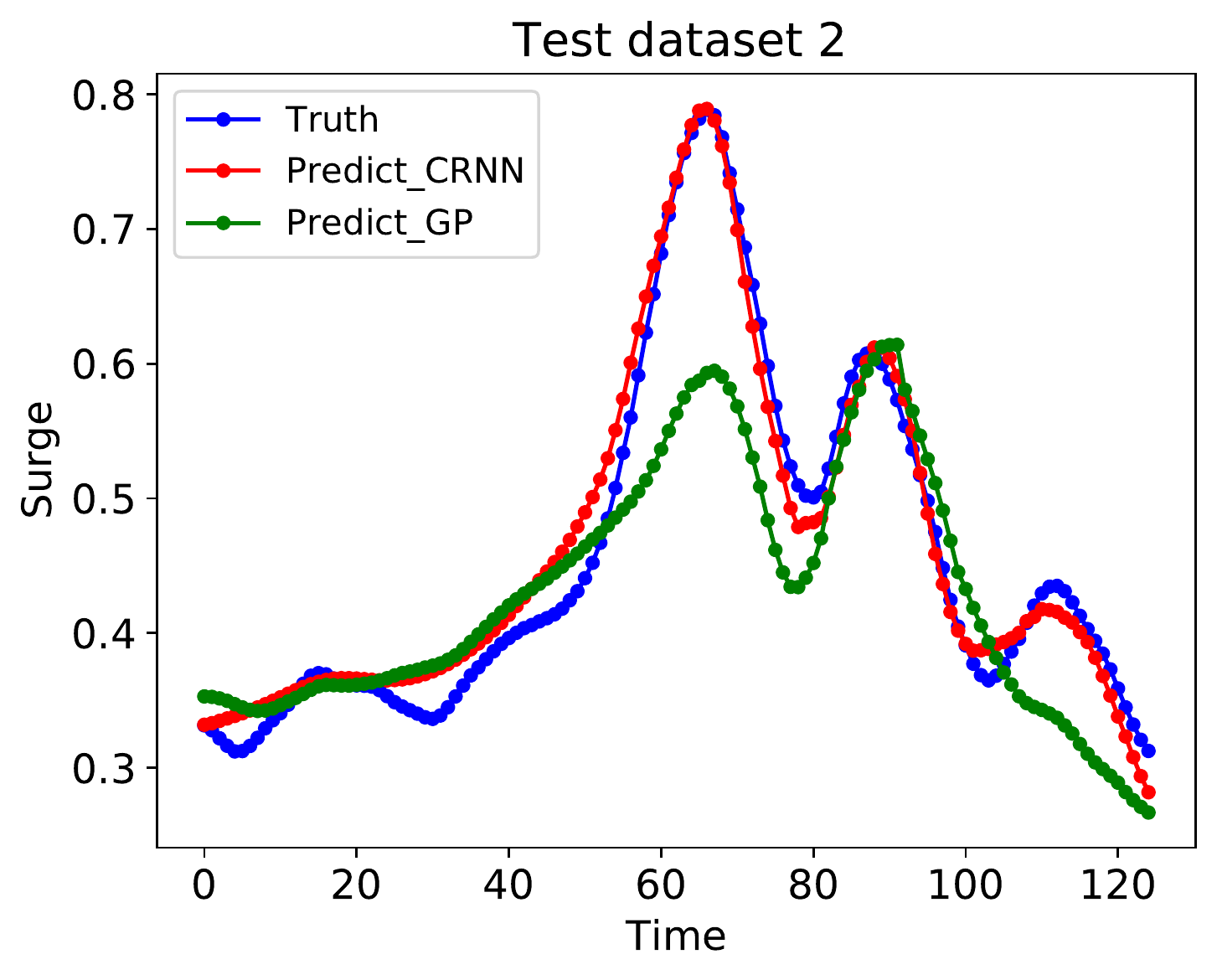} 
      \includegraphics[width=2.5in]{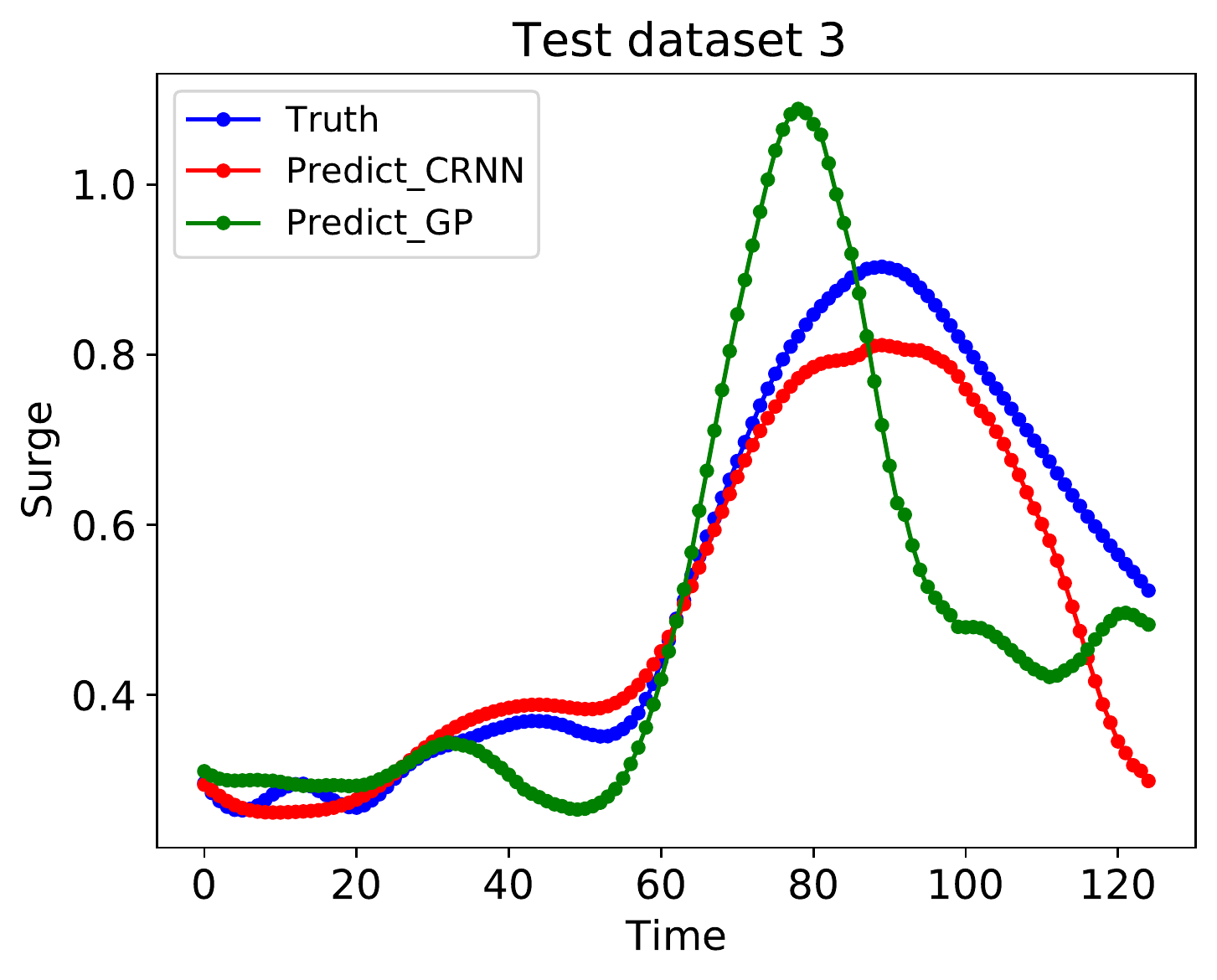} 
      \includegraphics[width=2.5in]{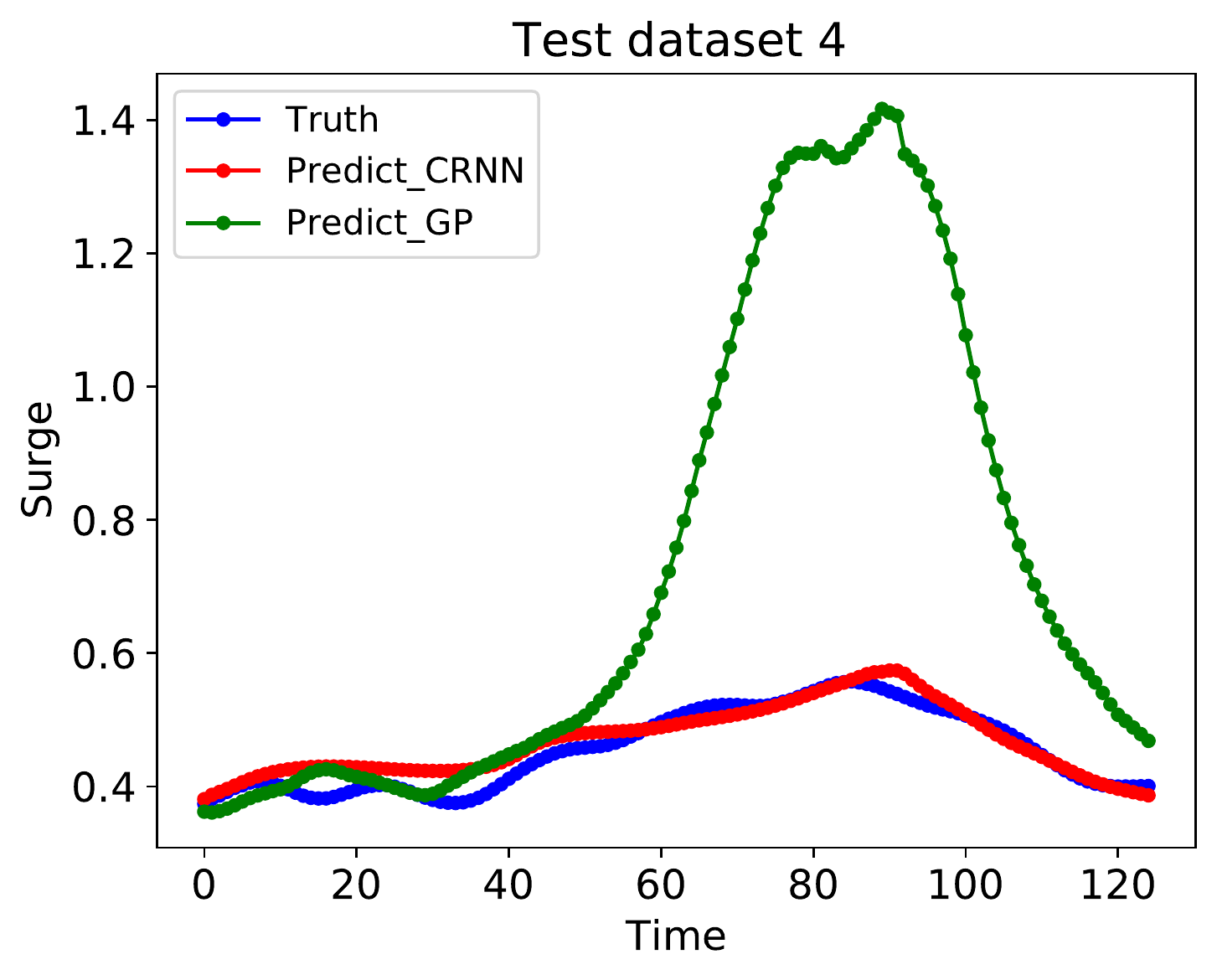} 
      \includegraphics[width=2.5in]{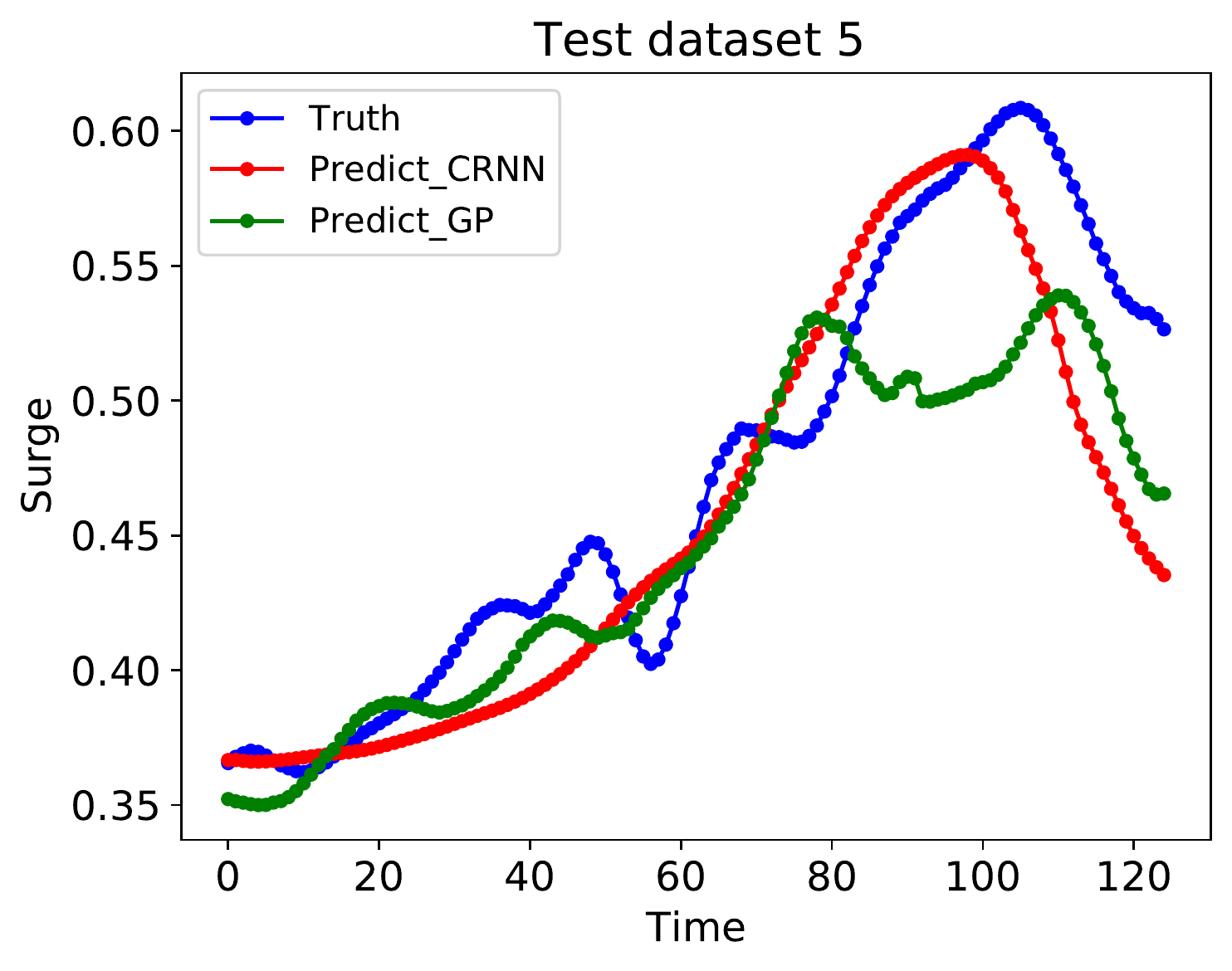} 
      \includegraphics[width=2.5in]{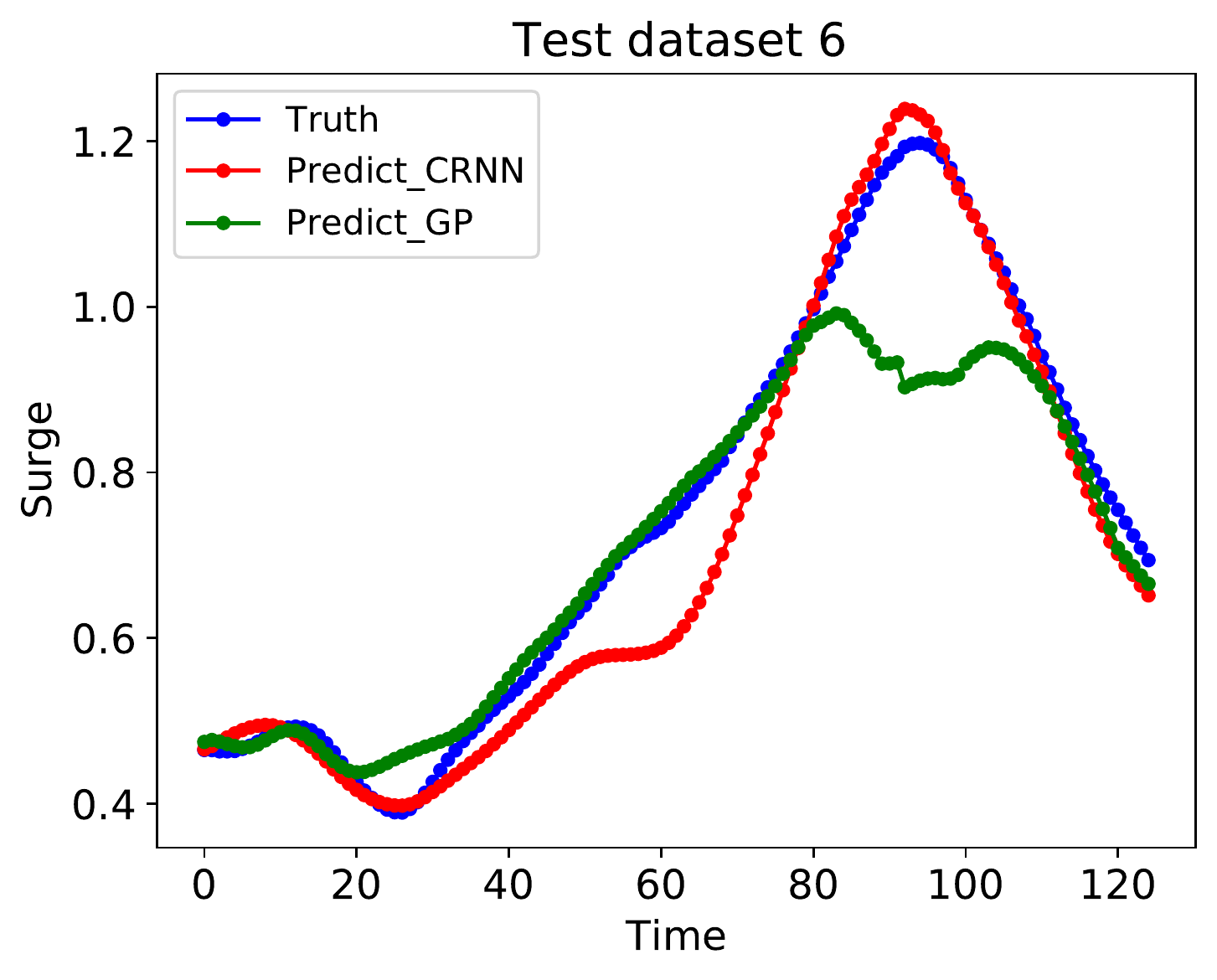} 
      \includegraphics[width=2.5in]{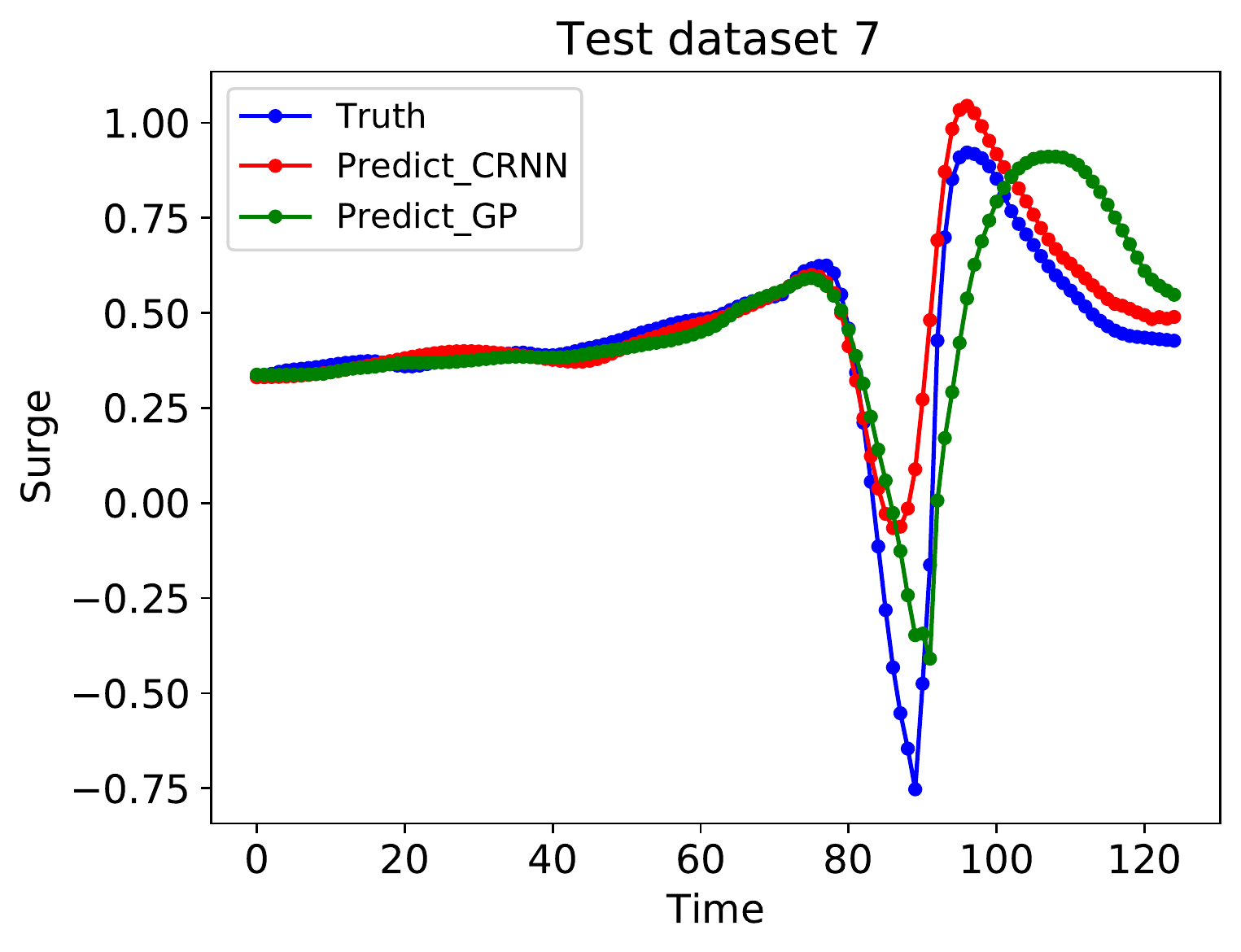} 
      \includegraphics[width=2.5in]{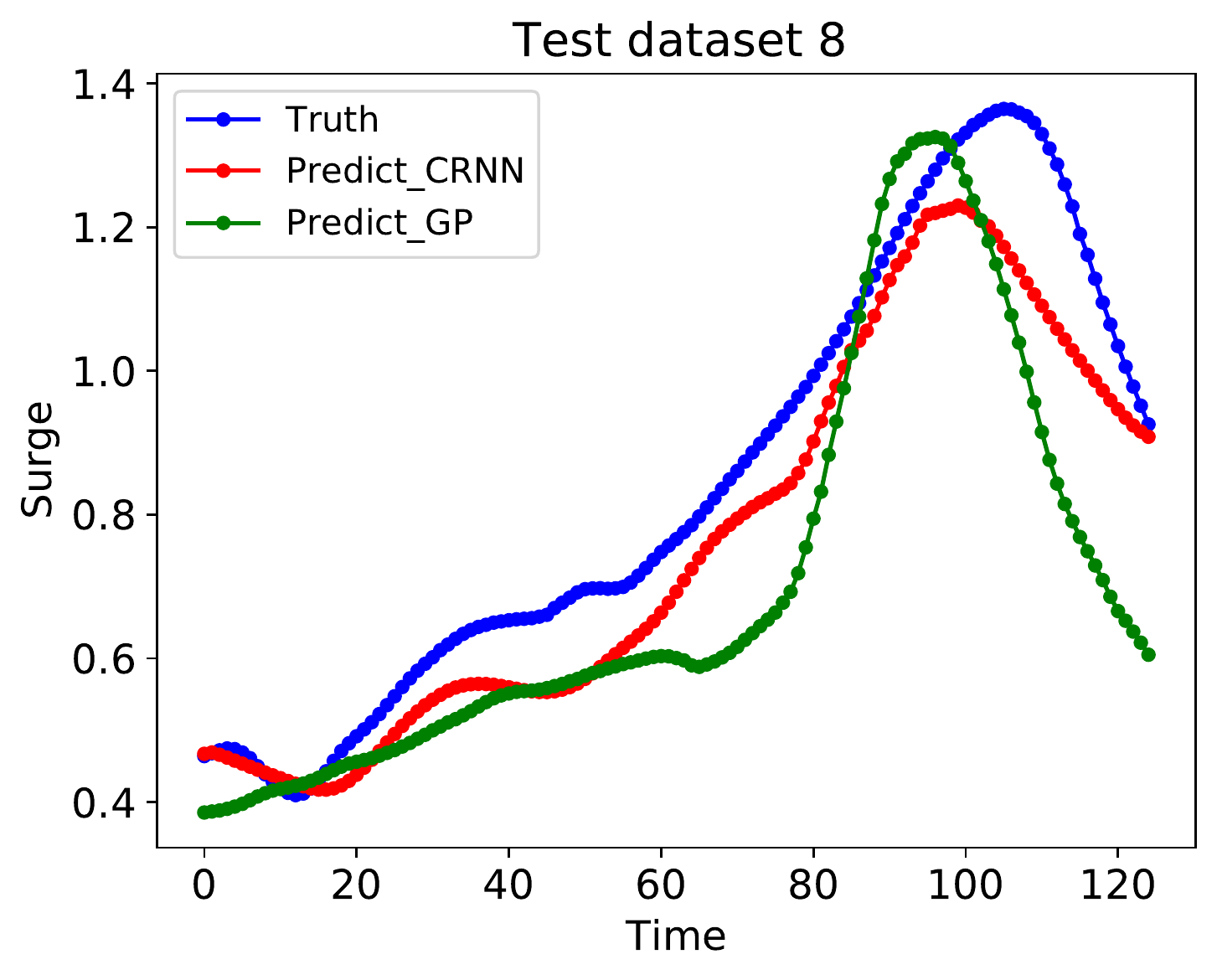}       
      \caption{\label{fig:CRNN_GP_early} Storm surge predictions for one grid SP in different test storms for a grid point in the coast front layers}
    \end{center}\hspace{2pc}%
\end{figure}

\begin{figure}[H]
    \begin{center}{}
      \includegraphics[width=2.5in]{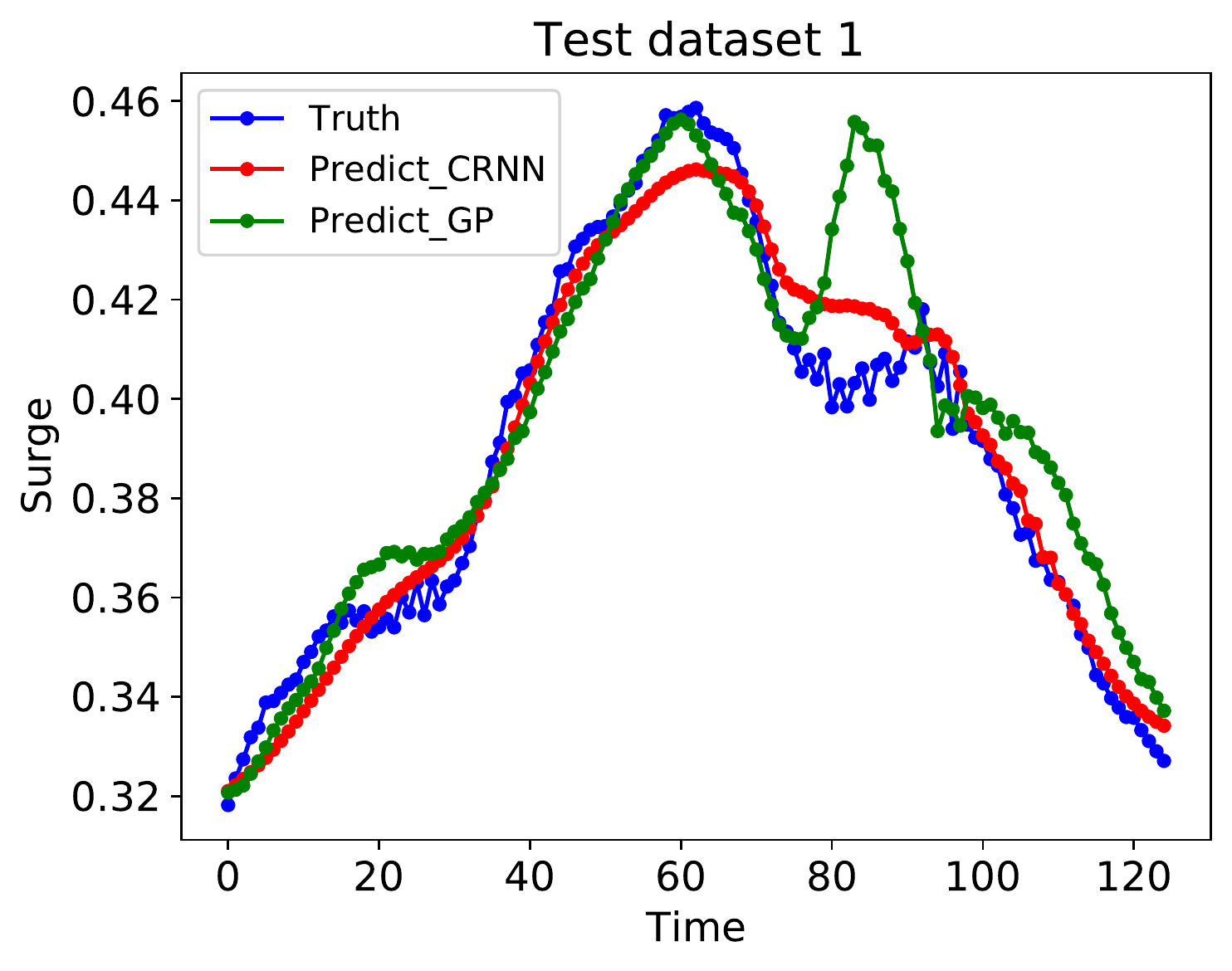}  
      \includegraphics[width=2.5in]{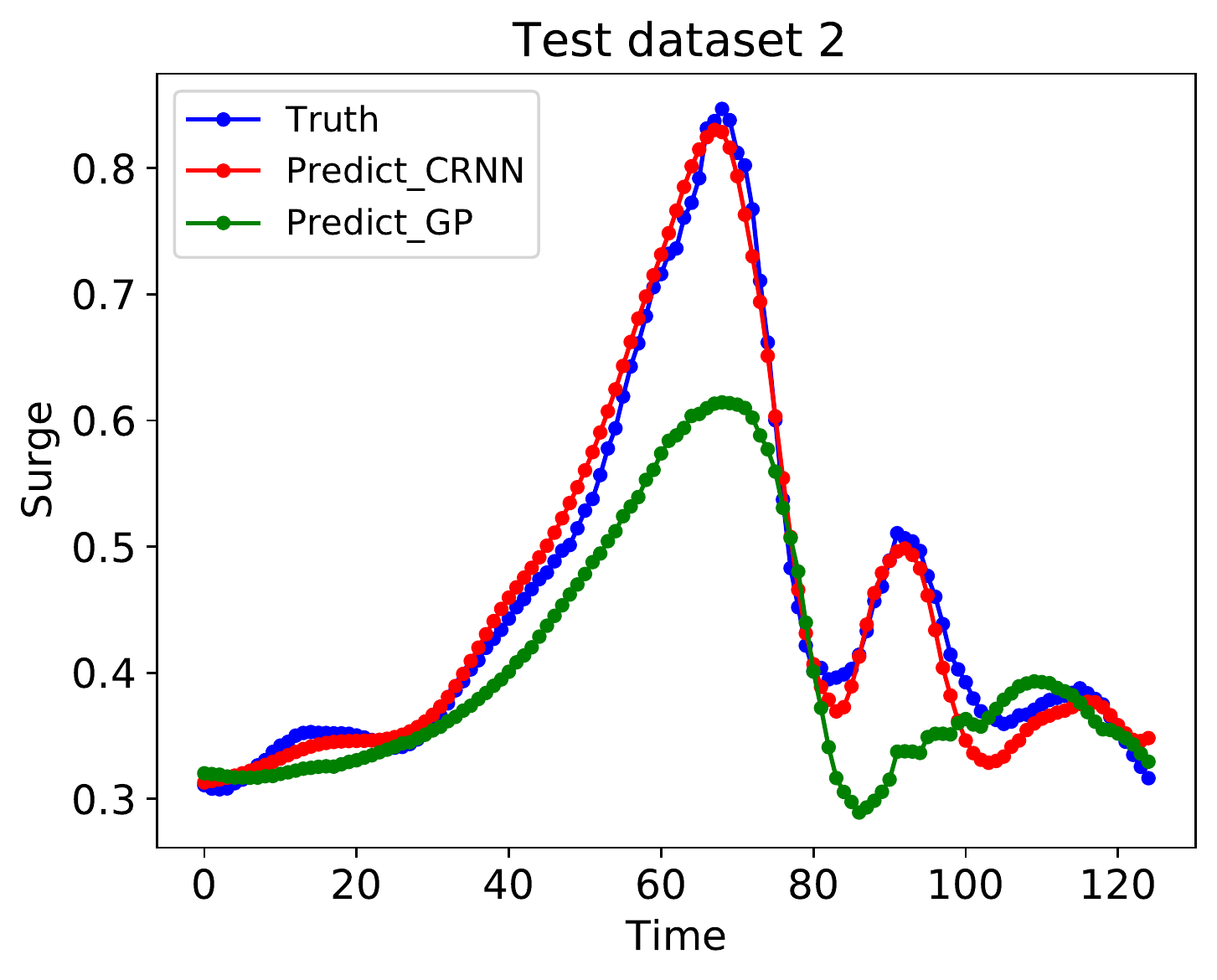} 
      \includegraphics[width=2.5in]{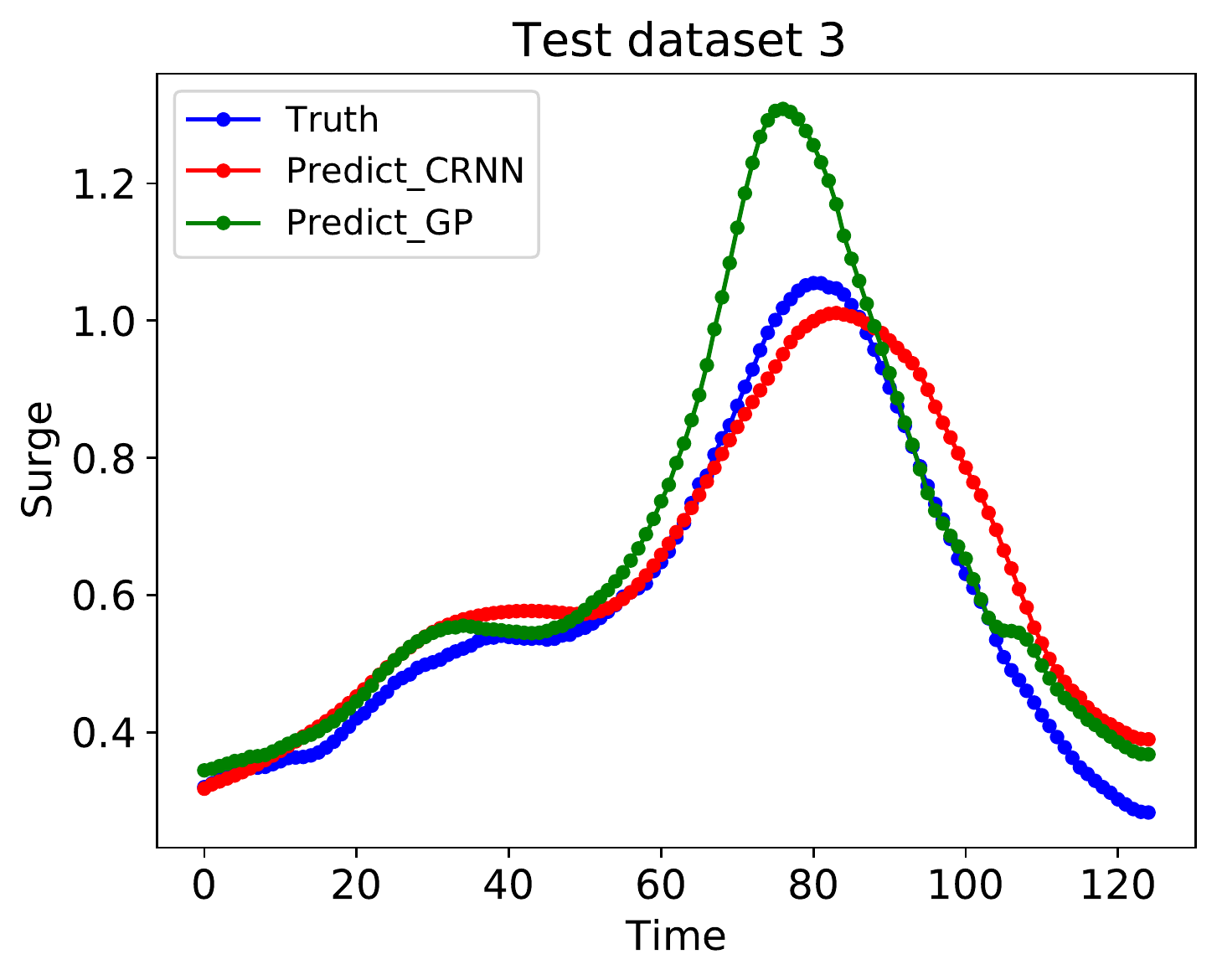} 
      \includegraphics[width=2.5in]{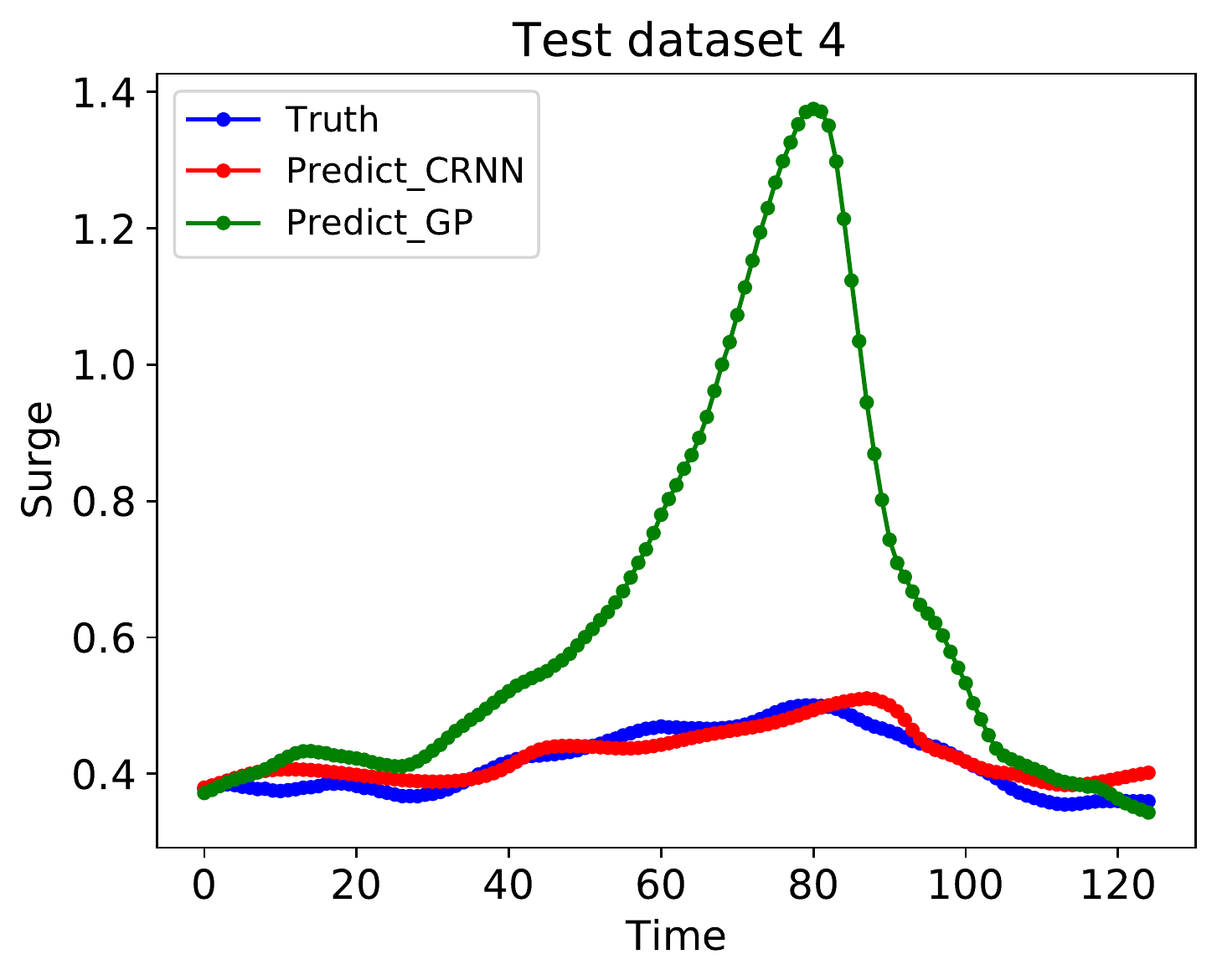} 
      \includegraphics[width=2.5in]{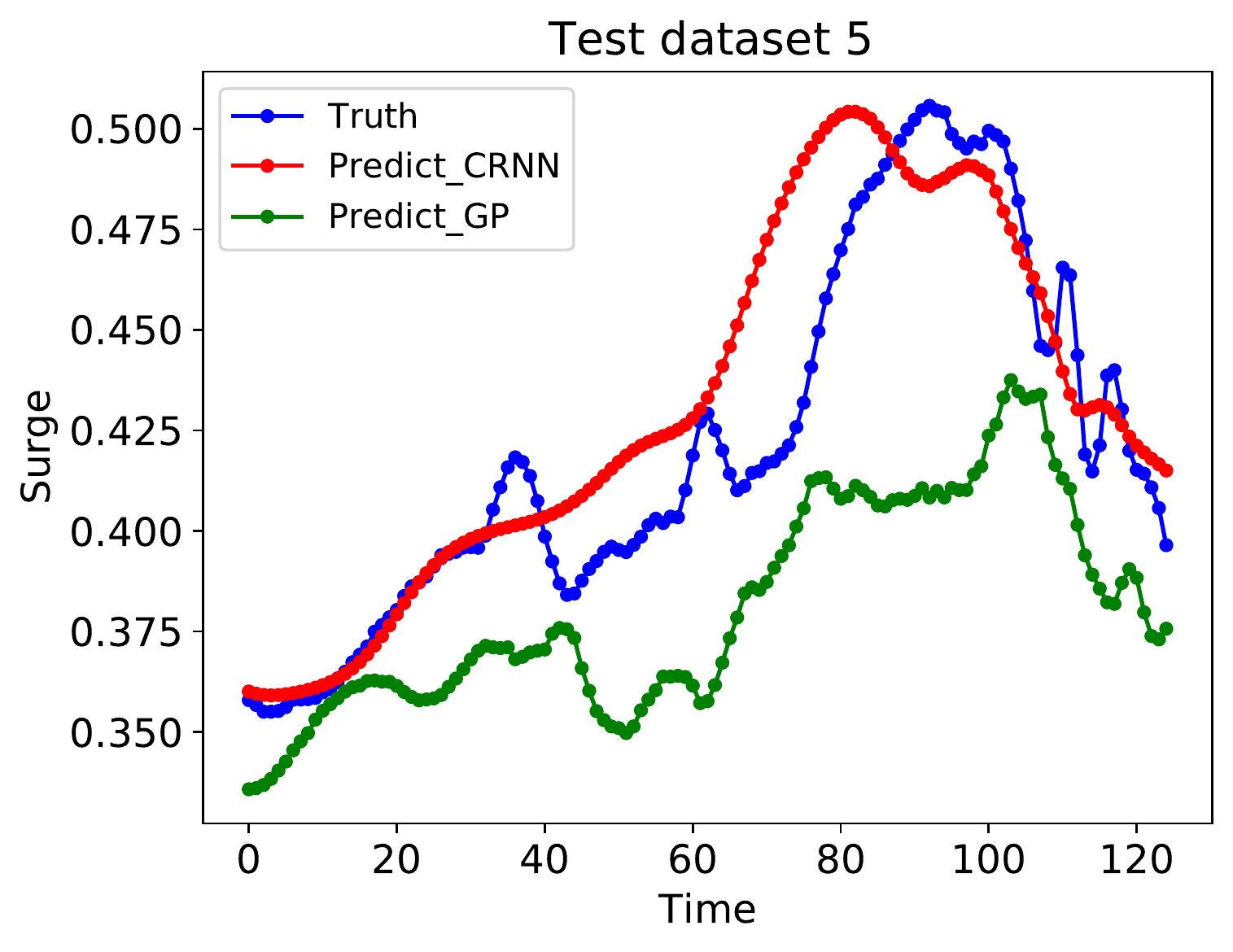} 
      \includegraphics[width=2.5in]{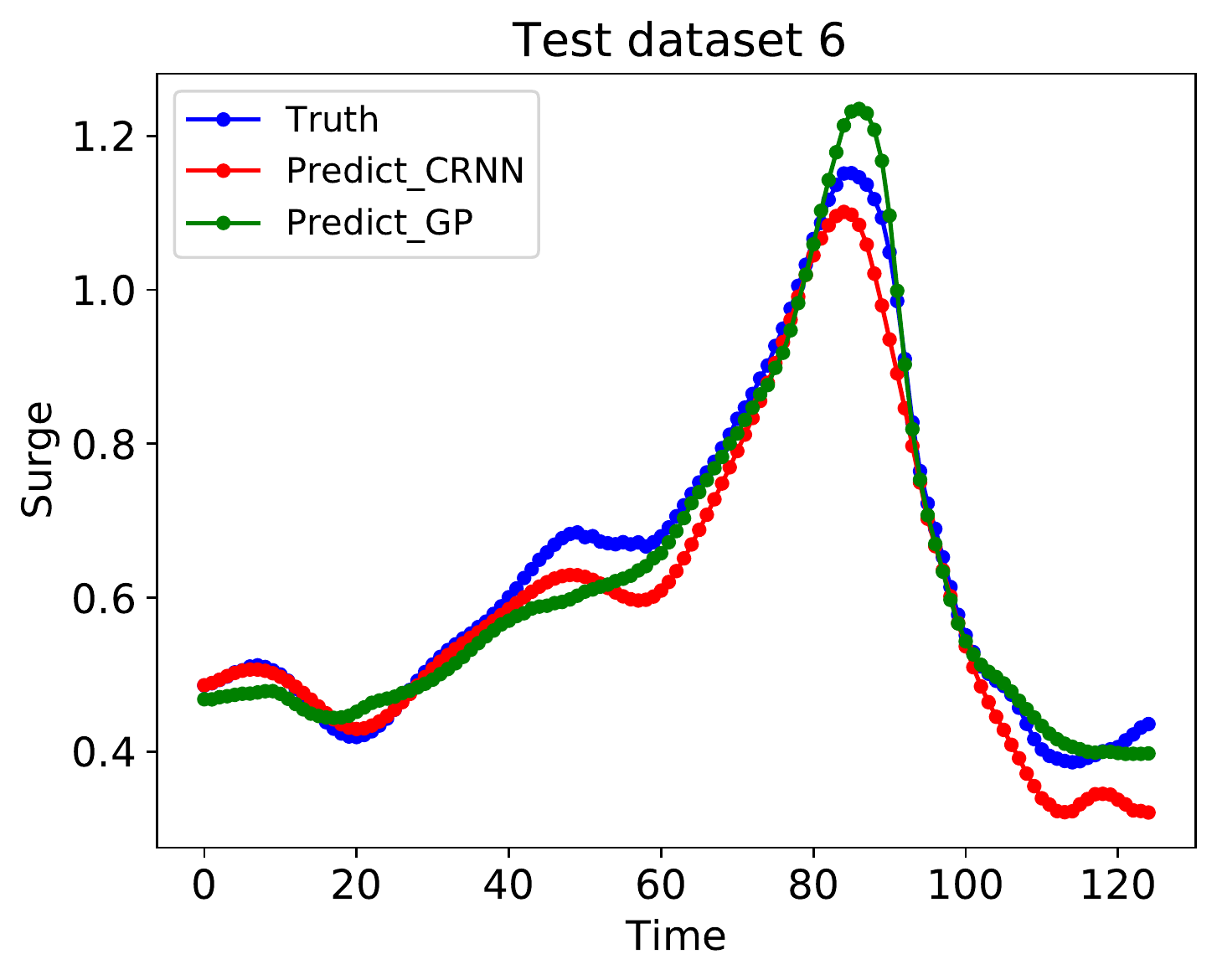} 
      \includegraphics[width=2.5in]{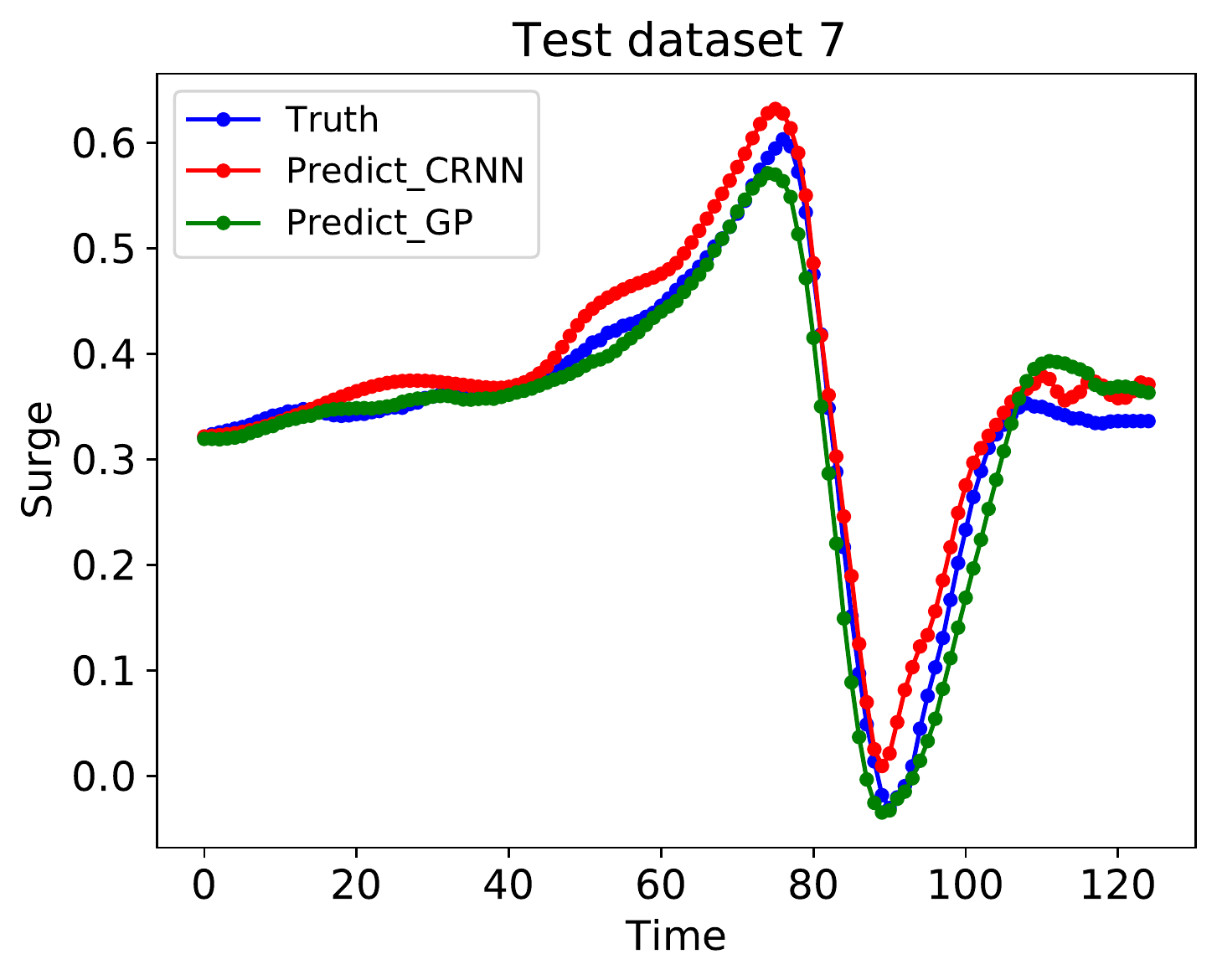} 
      \includegraphics[width=2.5in]{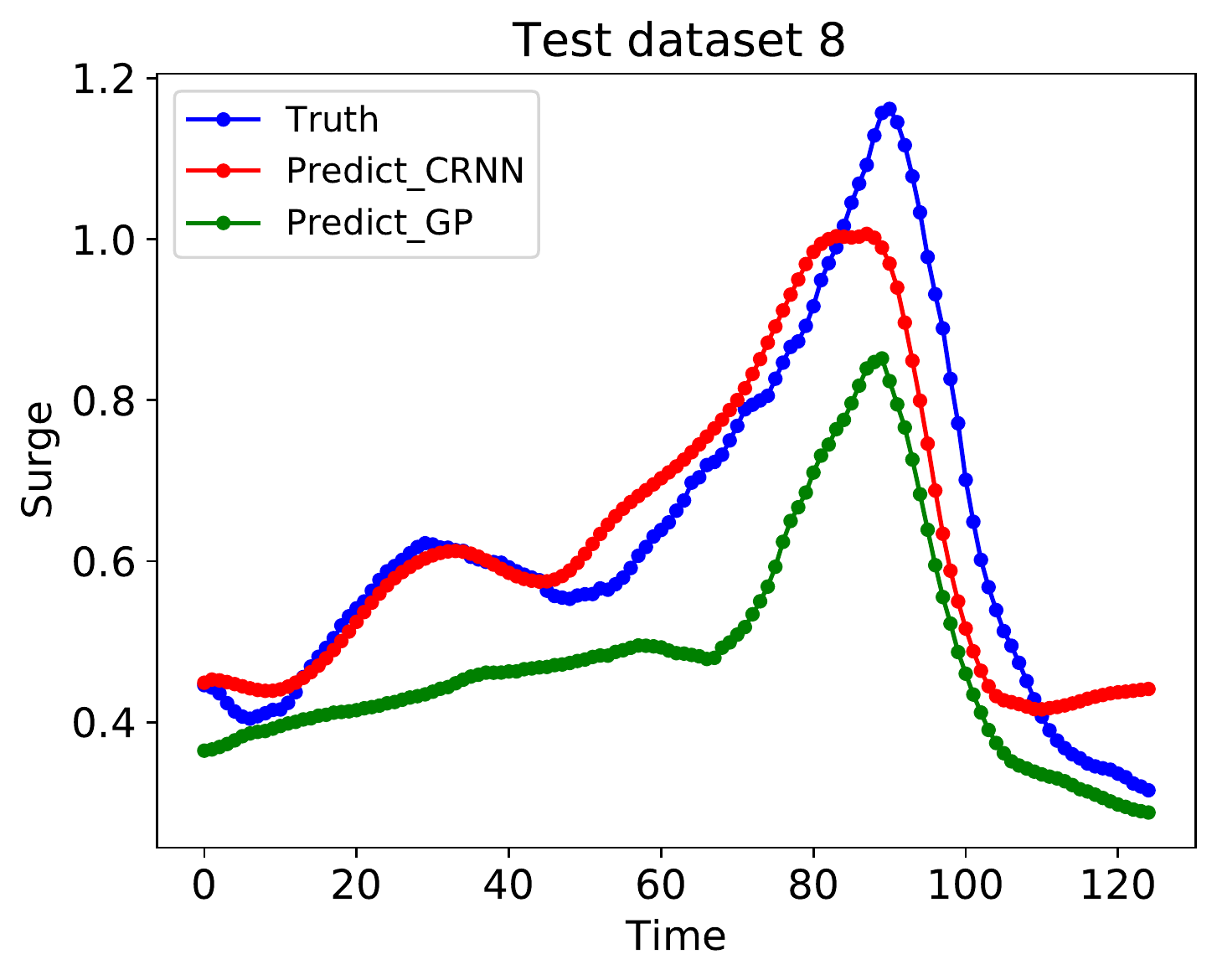}       
      \caption{\label{fig:CRNN_GP_end} Storm surge predictions for one grid SP in different test storms for a grid point in the coast back layers}
    \end{center}\hspace{2pc}%
\end{figure}

In this section, the comparison of the prediction by the developed convoloutional recurrent neural network and Gaussian process are shown in Figure \ref{fig:CRNN_GP_early} and Figure \ref{fig:CRNN_GP_end}. Figure \ref{fig:CRNN_GP_early} shows the comparison for a grid from the early layers of the coast and Figure \ref{fig:CRNN_GP_end} shows a grid from the last layer of grids in the coast. As it mentioned in Section \ref{section:Model Evaluation}, the results by convoloutioal recurrent neural network model are much better and more accurate than the Gaussian process. \\

\end{appendices}

\end{document}